\documentclass[final]{cvpr}

\usepackage{times}
\usepackage{epsfig}
\usepackage{graphicx}
\usepackage{amsmath}
\usepackage{amssymb}
\usepackage{microtype}

\usepackage[font={small}]{caption}
\usepackage{multirow}
\usepackage{booktabs}
\usepackage{bm}
\usepackage[dvipsnames]{xcolor}
\usepackage{import}

\usepackage[pagebackref=true,breaklinks=true,letterpaper=true,colorlinks,bookmarks=false]{hyperref}

\usepackage{cleveref}

\newcommand{\bx}{\bm{x}}

\newcommand{\bp}{\bm{p}}
\newcommand{\bq}{\bm{q}}
\newcommand{\bc}{\bm{c}}

\def\cvprPaperID{0000} %
\def\confYear{CVPR 2021}

\begin{document}

\title{KeypointDeformer: Unsupervised 3D Keypoint Discovery for Shape Control}

\author{%
Tomas Jakab$^{1,4*}$, Richard Tucker$^{4}$, Ameesh Makadia$^{4}$, Jiajun Wu$^{3}$, Noah Snavely$^{4}$, Angjoo Kanazawa$^{2, 4}$ \vspace{2pt}\\
 $^{1}$University of Oxford, $^{2}$UC Berkeley, $^{3}$Stanford University, $^{4}$Google Research\\}

\newenvironment{packed_enum}{
\begin{enumerate}
  \setlength{\itemsep}{1pt}
  \setlength{\parskip}{2pt}
  \setlength{\parsep}{0pt}
}{\end{enumerate}}

\newenvironment{packed_item}{
\begin{itemize}
  \setlength{\itemsep}{1pt}
  \setlength{\parskip}{2pt}
  \setlength{\parsep}{0pt}
}{\end{itemize}}

\newcommand{\tom}[1]{\textcolor{OliveGreen}{[\textbf{TJ says}: #1]}}
\newcommand{\todo}[1]{\textcolor{gray}{[\textbf{TODO}: #1]}}

\newcommand{\sznote}[1]{{\bf \color{red} [SZ\@: #1]}}
\newcommand{\jw}[1]{{\color{cyan} [JW\@: #1]}}
\newcommand{\richardt}[1]{{\bf\color{orange} [RT\@: #1]}}
\newcommand{\noah}[1]{{\color{orange} [NS\@: #1]}}
\newcommand{\ak}[1]{{\color{blue} [AK\@: #1]}}
\newcommand{\am}[1]{{\color{purple} [AM\@: #1]}}

\twocolumn[{
\renewcommand\twocolumn[1][]{#1}
\maketitle
    \centering
    \includegraphics[width=0.98\textwidth]{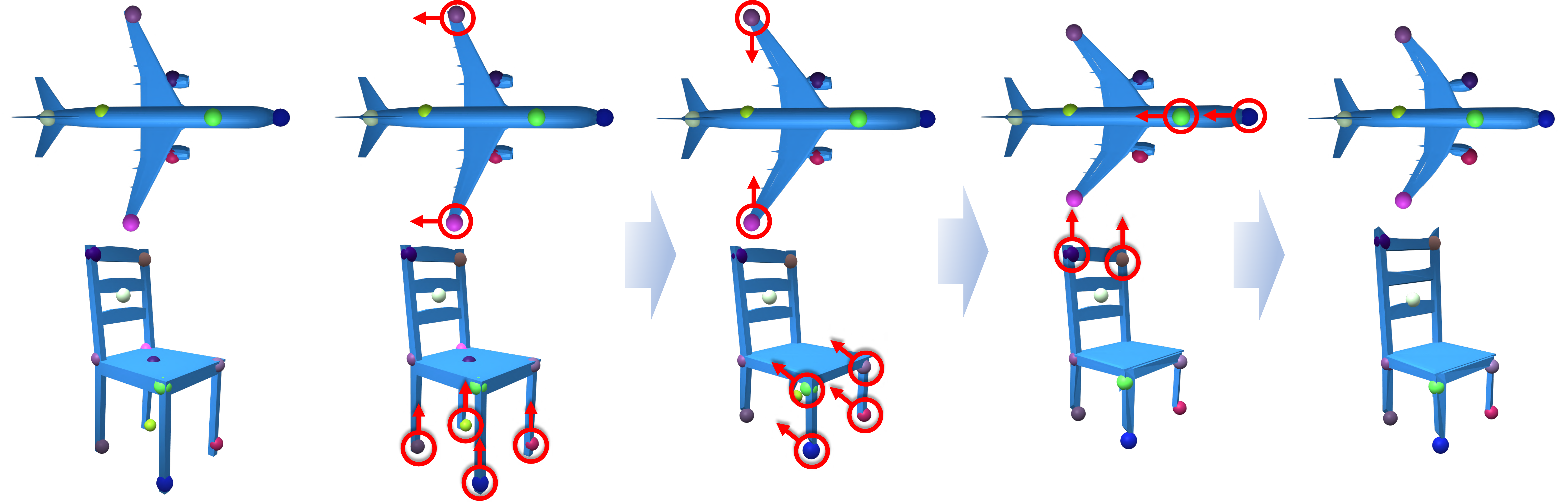}
   \captionof{figure}{
      \textbf{Controlling shape deformation with unsupervised 3D keypoints.}
      We discover unsupervised 3D keypoints that allow for intuitive control of an object's shape.
      This figure shows individual steps of interactive control.
      The red arrows illustrate the direction in which the keypoints are manipulated.
      Note that the resulting deformations are localized and object parts are deformed in an intuitive manner---\eg moving keypoints at the tip of the wings backward moves the wings backwards---all while preserving the details of the original shape.
      \label{fig:teaser}
   }
    \vspace{10pt}
}]

\begin{abstract}
   We introduce KeypointDeformer, a novel unsupervised method for shape control through automatically discovered 3D keypoints.
   We cast this as the problem of aligning a source 3D object to a target 3D object from the same object category.
   Our method analyzes the difference between the shapes of the two objects by comparing their latent representations.
   This latent representation is in the form of 3D keypoints that are learned in an unsupervised way.
   The difference between the 3D keypoints of the source and the target objects then informs the shape deformation algorithm that deforms the source object into the target object.
   The whole model is learned end-to-end and simultaneously discovers 3D keypoints while learning to use them for deforming object shapes.
   Our approach produces intuitive and semantically consistent control of shape deformations.
   Moreover, our discovered 3D keypoints are consistent across object category instances despite large shape variations.
   As our method is unsupervised, it can be readily deployed to new object categories without requiring annotations for 3D keypoints and deformations.
   Project page:~{\small \url{http://tomasjakab.github.io/KeypointDeformer}}.
\end{abstract}

\section{Introduction}
\let\thefootnote\relax\footnote{* Work done while interning at Google Research.}
Given the vast number of 3D shapes available on the Internet, providing users with intuitive and simple interfaces for semantically manipulating objects while preserving their key shape properties has a wide variety of applications in AI-assisted 3D content creation. In this paper, we propose to automatically discover intuitive and semantically meaningful control points for interactive editing, enabling detail-preserving shape deformation for object categories.
Specifically, we identify 3D keypoints as an intuitive and simple interface for shape editing. 
Keypoints are sparse 3D points that are semantically consistent across an object category.
We propose a learning framework for unsupervised discovery of such keypoints and a deformation model that uses the keypoints to deform a shape while preserving local shape detail.
We call our model \emph{KeypointDeformer}.

\Cref{fig:teaser} describes the inference-time use case of KeypointDeformer.
Given a novel shape, KeypointDeformer predicts 3D keypoints on the surface. 
If a user manipulates a keypoint on a chair leg upwards, the entire leg is deformed in the same direction (bottom). 
Our approach optionally enables the use of a categorical deformation prior on these edits, such that if a user moves one side of an airplane wing backwards, the opposite side of the wing is deformed symmetrically in the same direction (top)---while if the user wishes to only move one side of the wing, our approach also allows this. 
Our framework enables stand-alone shape edits or shape alignment between two shapes, and can also synthesize novel variations of shapes for amplifying stock datasets. 

While 3D keypoints may be a good proxy for shape editing, obtaining explicit supervision for keypoints and deformation models is not only expensive but also ill-defined. As such, we propose an unsupervised framework for jointly discovering the keypoints and the deformation model. To solve our problem, we devise two components that operate in concert: (1) a method for discovering and detecting keypoints, and (2) a deformation model that propagates keypoint displacements to the rest of the shape. To achieve these, we set up a proxy 
learning task where the goal is to align a source shape with a target shape, where the two can represent very different instances of a category.
We also propose a simple yet effective keypoint regularizer that encourages learning of semantically consistent keypoints that are well-distributed, lie close to the object surface and implicitly preserve underlying shape symmetries.
The result of our training approach is a deformation model that deforms a shape based on automatically discovered 3D control keypoints. Since the keypoints are low-dimensional, we can further learn a category prior on these keypoints, enabling semantic shape editing from sparse 
user inputs.

Overall, our method has following key benefits:
\begin{packed_enum}
    \item It gives users an intuitive and simple way to interactively control object shapes.
    \item Both the keypoint prediction and deformation model are unsupervised.
    \item We show that keypoints discovered by our method are better for shape control than other kinds of keypoints, including manually annotated ones.
    \item Our unsupervised 3D keypoints are semantically consistent across object instances of the same category giving us sparse correspondences.
\end{packed_enum}
We evaluate the semantic consistency of our unsupervised 3D keypoints on standard benchmarks,
and achieve state-of-the-art results among unsupervised methods.
We also demonstrate the suitability of our keypoints for shape deformation.
Finally, we provide qualitative results of user-guided interactive shape control, and include videos of interactive shape control on our project page.

\section{Related Work}
\begin{figure*}
\centering
\includegraphics[width=0.95\textwidth]{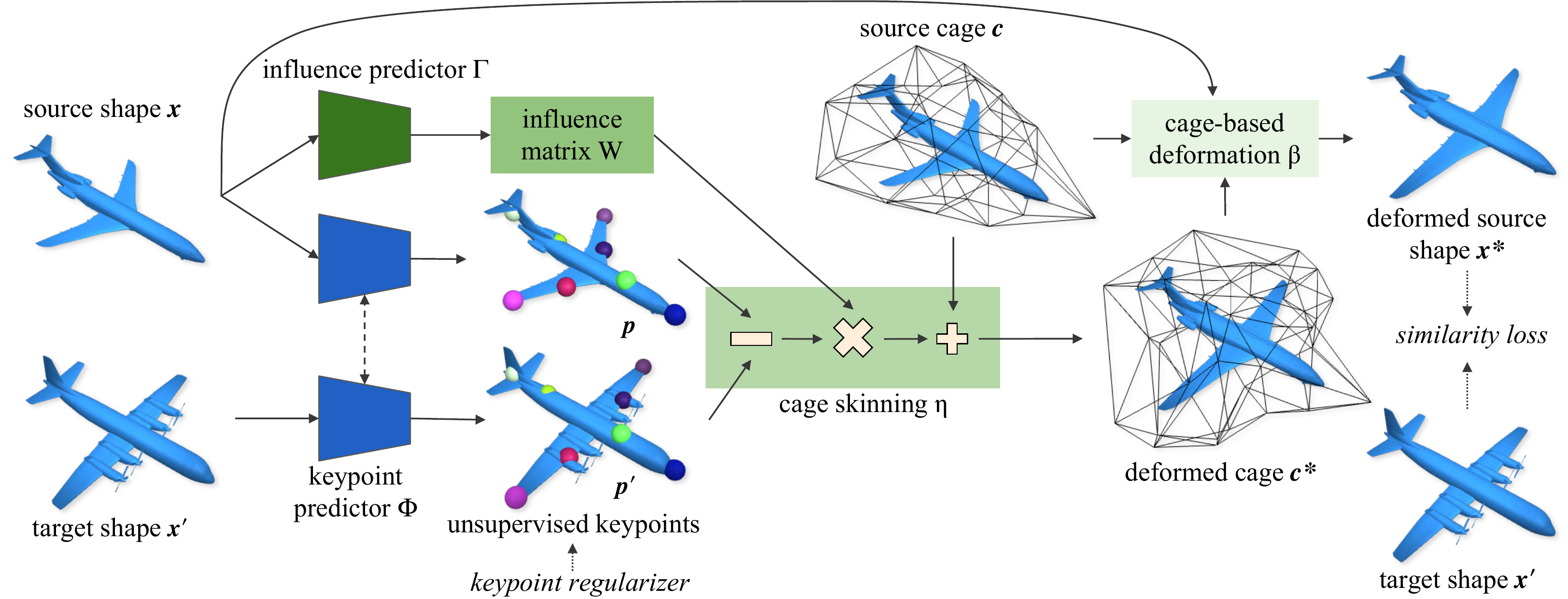}
\caption{
    \textbf{Model.}
    Our model aligns the source shape $\bx$ with the target shape $\bx'$ using predicted unsupervised keypoints $\bp$ and $\bp'$.
    The unsupervised keypoints describe the objects pose and work as control points for the deformation.
    The model is trained end-to-end using a similarity loss between the deformed source shape $\bx^*$  and the target shape $\bx'$, as well as a keypoint regularization loss.
    During interactive shape manipulation at test time, a user can choose to input only the source shape $\bx$ that the keypoint predictor $\Phi$ uses to estimate a set of unsupervised keypoints $\bp$.
    The user can then manually control the keypoints $\bp$ obtaining $\bp'$ target keypoints that are fed into the deformation model to produce the deformed source shape $\bx^*$ as demonstrated in~\Cref{fig:teaser},~\Cref{f:control} and in the supplementary videos on our project page.
}
\label{f:model}
\end{figure*}

\noindent \textbf{Shape deformation.}
Our approach is closely related to 
detail-preserving deformations studied in geometric modeling, including Laplacian-based shape editing~\cite{sorkine2006differential}, As-Rigid-As-Possible shape deformation~\cite{sorkine2007rigid}, and cages~\cite{ju2005mean}. 
While these approaches enable shape editing via many forms of user-specified constraints (e.g., points or sets in an optimization framework), a major challenge is that they rely purely on geometric properties and do not consider semantic attributes or category-specific shape priors for deformation. 
Such priors can be obtained from artists painting the object surface with stiffness properties~\cite{botsch2006primo} or learned from a set of meshes with known 
correspondence~\cite{popa2006material}. However, such supervisions are prohibitively expensive to obtain and are not applicable to novel shapes. %
Yumer~\etal\cite{yumer2015semantic} address this issue in a data-driven framework that provides a set of sliders that control the attributes of a given shape. However, this approach requires a set of predefined attributes obtained from expert annotations. We propose an unsupervised approach, and provide users with direct semantic deformation handles in the form of keypoints. Furthermore, our formulation can incorporate a category-specific deformation basis on the discovered 3D keypoints,  allowing for semantically consistent user edits from sparse 
keypoints edits (such that if one side of an airplane wing is extended, the other opposite side also extends). 

Another related problem is deformation transfer~\cite{sumner2004deformation}, which transfers the deformation exhibited by a source mesh onto a target mesh via known correspondences between shapes.
Recent approaches employ deep learning to implicitly learn the shape correspondences to align two shapes~\cite{yifan2020neural,hanocka2018alignet,wang20193dn}. While we also use a shape alignment objective to train our framework, we make our intermediate control explicit in the form of keypoints, which allows for stand-alone shape editing. In contrast, prior approaches always require a target shape to express the desired deformation.
\medskip
\noindent \textbf{User-guided shape editing.}
Our approach is related to recent deep learning--based methods that learn generative models of shapes for interactive editing. 
Tulsiani \etal~\cite{tulsiani2017learning} learn to abstract shapes in terms of primitives, which can be used to edit the shape by transferring primitive deformations to the surface. 
However, shape editing is not their primary focus, and it is unclear how well the direct transfer of primitive transformations preserve local shape detail. 
Recent approaches take this idea further by learning a generative model of primitives in the form of set of point-based primitives~\cite{hao2020dualsdf}, shape handles~\cite{Gadelha_2020_CVPR}, or disconnected shape manifolds~\cite{mehr2019disconet}. 
These methods enable interactive editing by searching for latent primitive representations that best match user edits. 
However, they require an involved user interface via sketching or directly manipulating the underlying set of primitives.
Most critically, as the edits are based on generative models, these approaches may change the local details of the original shape.
In contrast, we directly deform the source shape, leading to better preservation of shape detail.
We qualitatively compare our approach to DualSDF~\cite{hao2020dualsdf} to illustrate this benefit.  

\medskip
\noindent \textbf{Unsupervised keypoints.}
While the problem of unsupervised keypoint discovery is well studied in 2D~\cite{thewlis2017unsupervised,zhang2018unsupervised,koepke2018self,jakab2018unsupervised,thewlis2019unsupervised,jakab2020self}, this problem is relatively under-explored in 3D. Suwajanakorn \etal~\cite{suwajanakorn2018discovery} detect 3D keypoints from a single image using 3D pose information as supervision. Here we focus on learning 3D keypoints on 3D shapes. Chen \etal~\cite{chen2020unsupervised} output a structured 3D representation to obtain sparse or dense shape correspondences. Closest to our approach in terms of 3D keypoint discovery is that of Fernandez \etal~\cite{fernandez2020unsupervised}, which impose explicit symmetric constraints. In this work, we discover unsupervised keypoints for the purpose of shape control. While we focus on shape editing, our formulation results in state-of-the-art 3D keypoints for semantic consistency. Such unsupervised keypoints may be useful for robotics applications that use 3D keypoints as a latent representation for control~\cite{manuelli2019kpam,gao2019kpam}, and which currently require manually defined 3D keypoints as supervision.

\section{Method}

Our aim is to learn a keypoint predictor $\Phi: \bx \rightarrow \bp$ that maps a 3D object shape $\bx$ to a sparse set of semantically consistent 
 3D keypoints $\bp$. 
We also want to learn a conditional deformation model on keypoints $\Psi: (\bx, \bp, \bp') \rightarrow \bx'$ that deforms the shape $\bx$ in accordance to the deformed control keypoints, where $\bp$ describes the initial (source) keypoint locations and $\bp'$ the target locations. %
Obtaining explicit supervision for keypoints and the deformation model is expensive and ill-defined. As such, we propose an unsupervised learning framework for training these functions.
We do so by designing an auxiliary task of pair-wise shape alignment, where the key idea is to jointly learn keypoints and a deformation model that can bring 
two random shapes into alignment. Specifically, our model 
first predicts keypoint locations on the source and target shapes using a Siamese network.
We then deform the source shape according to the correspondence provided by the discovered keypoints. 
In order to preserve local shape detail, we employ a cage-based deformation method, conditioned on keypoints.
We devise a novel and highly effective, yet simple, keypoint regularization term that encourages keypoints to be well-distributed and lie close to the object surface. 
\Cref{f:model} provides a schematic illustration of our framework. %

\subsection{Shape Deformation with Keypoints}

We first predict keypoints from source and target meshes by representing each object as a point cloud $\bx \in \mathbb{R}^{3\times N}$, uniformly sampled from the object mesh. The keypoint predictor $\Phi$ takes the shape as an input $\bx$ and outputs an ordered set of 3D keypoints $\bp = (p_1,\dots,p_K)\in \mathbb{R}^{3\times K}$.
The encoder is shared for both the source and target in a Siamese architecture. %
The shape deformation function $\Psi$ takes the source shape $\bx$ represented as a point cloud $\bx$
 as well as source keypoints $\bp$ and target keypoints $\bp'$.
The keypoints $\bp$ and $\bp'$ are estimated by the keypoint predictor $\Phi$.
At test time, the user can input their own target keypoints $\bp'$ for interactive shape deformation as illustrated in \Cref{f:model}.

In order to deform the object shape in a manner that preserves its local shape detail, we use the recently introduced differentiable cage-based deformation algorithm~\cite{yifan2020neural}.
Cages are a classical shape modeling method~\cite{ju2005mean,joshi2007harmonic,lipman2008green} that use a coarse enclosing mesh that is associated with the shape.
Deforming the cage mesh results in an interpolated deformation of the enclosed shape.
The cage-based deformation function $\beta: (\bx, \bc, \bc^*) \rightarrow \bx^*$ takes a source control cage $\bc$ and a deformed control cage $\bc^*$, and deforms the input shape $\bx$ that is in the form of a mesh or a point cloud.
We automatically obtain the source cage $\bc$ for each shape by starting with a spherical shape that is larger than the source shape $\bx$ and iteratively pulling each of the cage vertices $\bc_V$ towards the centre of the object until it is within a small distance from the object surface. The resulting cages are illustrated in \Cref{f:model}. While cages are a reliable method for shape-preserving deformation, modifying cages to achieve a desired deformation is not necessarily intuitive, particularly to novice users, because the cage vertices do not lie on the surface, do not have a coarse structure, and are not semantically consistent across different shapes. We propose keypoints as an intuitive handle to manipulate the cages.

In order to control the object deformation using our discovered keypoints, we need to associate them with the cage vertices.
We do so with a linear skinning function that takes the relative differences between the source and target keypoints $\Delta \bp = \bp' - \bp$ and associates each of them with the source cage vertices $\bc_V$ using an influence matrix $W \in \mathbb{R}^{C \times K}$ that we learn in an end-to-end manner, where $C$ is the number of cage vertices and $K$ is the number of discovered keypoints.
The resulting deformed cage vertices $\bc^*_V$ are then defined as 
\begin{equation}
 \bc^*_V = \bc_V + W \Delta \bp.   
\end{equation}

In order to adjust for the fact that cages are unique to each shape, we represent the influence matrix as a function of the input shape $\bx$.
Specifically, the influence matrix is a composition $W(\bx) = W_C + W_I(\bx)$ of a canonical $W_C$ matrix that is shared with all instances of the object category and an instance specific offset $W_I$ that is predicted from the source shape $\bx$ using an influence predictor $W_I = \Gamma(\bx)$.
We regularize the instance specific $W_I$ matrix by minimizing its Frobenius norm to prevent overfitting of the resulting influence matrix $W$.
We denoted this regularizer as $\mathcal{L}_\text{inf}$.
Finally, we limit the matrix $W$ to only influence at most $M$ nearest cage vertices per each keypoint to encourage locality. 

\begin{figure}[t]
  \small
  \centering
  \includegraphics[width=1\linewidth]{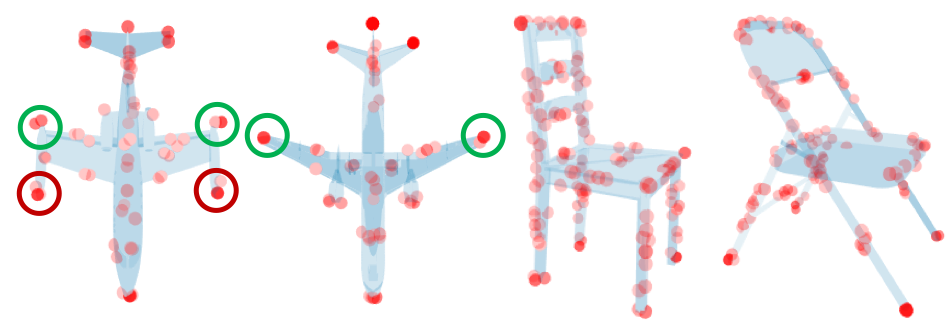}
  (a) frequency of sampled regularizer points
  \includegraphics[width=1\linewidth]{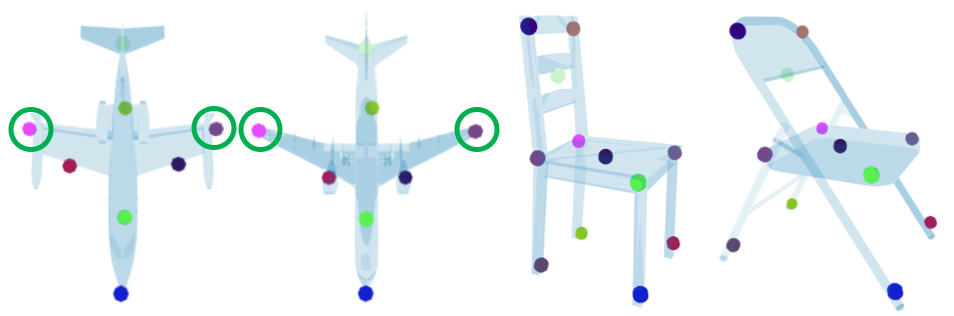}
  (b) unsupervised keypoints predictions
  \caption{
      \textbf{Farthest Point Keypoint regularizer.}
      We use farthest point sampling with a random starting point to regularize the predicted keypoints.
      \textbf{(a)} illustrates the frequency of a given point being sampled by the farthest point sampling algorithm.
      Darker colours indicate higher probability of a point being sampled.
      The expected locations of sampled points provide good coverage and inherently follow the symmetry of the original shape.
      Also, a subset of them tend to be semantically stable across different object instances.
      Using expected sample locations as a prior for keypoint location works well as the keypoint predictor will learn to be robust to noise in these sampled points.
      This can be seen in the example of the airplane where the tips on the fuel tanks (shown in \textbf{\color{BrickRed}red} circle) are ignored, and the keypoints are instead predicted \textbf{(b)} at the wingtip (shown in \textbf{\color{OliveGreen}green} circle) location that is more consistent across the dataset (most planes have wings, but many lack fuel tanks).
  }
  \label{f:init_points}
\end{figure}

\subsection{Losses and Regularizers}
Our KeypointDeformer is trained end-to-end with stochastic gradient descent by minimizing a similarity loss between the source and target shape, as well as a keypoint regularization term and instance-specific influence matrix regularization term.

\medskip
\noindent \textbf{Similarity loss.}
Ideally, we would like to compute the similarity between the deformed source shape $\bx$ and the target shape $\bx'$ using known correspondences between the meshes.
However, such correspondence is not available since we aim to train on generic collections of object category CAD models.
We approximate the similarity loss by computing the Chamfer distance between the deformed source $\bx^*$ and the target shape $\bx'$ represented as point clouds.
We denote this loss as $\mathcal{L}_\text{sim}$.

\medskip
\noindent \textbf{Farthest Point Keypoint regularizer.}
We propose a simple, yet highly effective keypoint regularizer $\mathcal{L}_\text{kpt}$ that encourages predicted keypoints $\bp$ to be well-distributed, lie on the object surface, and preserve the symmetric structure of the underlying shape category.
Specifically, we devise a Farthest Sampling Algorithm to sample an unordered set of points $\bq = \{q_1,\dots,q_J\}\in \mathbb{R}^{3\times J}$ from the input shape $\bx$ represented as a point cloud.
The initial point for sampling is chosen at random, and hence each time we compute this regularization loss a different set of sampled points $\bq$ is used.
Given these stochastic farthest points, the regularizer minimizes the Chamfer distance between the predicted keypoints $\bp$ and sampled points $\bq$.
In other words, the regularizer encourages the keypoint predictor $\Phi$ to place the discovered keypoints $\bp$ at the expectation of the sampled farthest points $\bq$.
\Cref{f:init_points} illustrates the properties of the sampled regularizer points.
The sampled points provide equally spaced coverage of the input object shape $\bx$, are relatively stable across different instances, and preserve the symmetric structure of the original input shapes.

Another intuition behind this regularization is that we can consider the sampled farthest points $\bq$ as a noisy prior over keypoint locations.
This prior is not perfect---it may miss important points in some models, or place spurious points in others---but the neural network keypoint predictor will learn keypoints in a way that is robust to such noise, and instead, prefer to predict keypoints at consistent locations, as demonstrated in~\Cref{f:init_points}.

\medskip
\noindent \textbf{Full objective.}
In summary, our full training objective is
\begin{equation}
    \mathcal{L} = \mathcal{L}_\text{sim} + \alpha_\text{kpt} \mathcal{L}_\text{kpt} + \alpha_\text{inf} \mathcal{L}_\text{inf}
\end{equation}
where $\alpha_\text{kpt}$ and $\alpha_\text{inf}$ are scalar loss coefficients. %
Our method is simple and does not require additional shape specific regularization for shape deformation, such as the point-to-surface distance, normal consistency, and symmetry losses employed in~\cite{yifan2020neural}. %
This is due to the fact that keypoints provide a low-dimensional correspondence between shapes and that cage deformations are a linear function of these keypoints, preventing extreme deformations that result in unwanted local shape deformations. 

\subsection{Categorical Shape Prior}\label{s:prior}
Since we represent an object shape as a set of semantically consistent keypoints, we can obtain a categorical shape prior by computing PCA on the keypoints predicted on the training set.
This prior can be used to guide keypoint manipulation.
For example, if user edits a single keypoint on an airplane wing, the remaining keypoints can be ``synchronized'' according to a prior by finding the PCA basis coefficients that best reconstruct the new position of the edited keypoint.
The resulting reconstructed set of keypoints follow the prior defined by the data. This prior also allows sampling of novel shapes via sampling a new set of keypoints.
This set of keypoints can be then used to deform the shape using our deformation model in order to, for instance, automatically augment libraries of stock 3D models.

\section{Experiments}
The main objectives of our experiments are to evaluate whether (1) our discovered keypoints are in general of good quality as keypoints (\Cref{s:semantic}), (2) our discovered keypoints are better suited for shape deformation than other keypoints (\Cref{s:deform}), and (3) our method allows for intuitive shape control (\Cref{s:control}).
The supplementary material contains extended version of results and ablation studies.
\begin{table}
  \small
  \centering

      \setlength{\tabcolsep}{4pt}
      \begin{tabular}{l|ccccccc}
        \toprule
         & airplane & car & chair & motorbike & table \\ \midrule
          Chen~\etal~\cite{chen2020unsupervised} & 0.69 & 0.39 & 0.78 & 0.91 & 0.75 \\
          Fernandez~\etal~\cite{fernandez2020unsupervised} & 0.78 & 0.66 & 0.80 & 0.90 & 0.85 \\
          ours & \textbf{0.85} & \textbf{0.73} & \textbf{0.88} & \textbf{0.93} & \textbf{0.92} \\
        \bottomrule
      \end{tabular}

  \caption{
      \textbf{Semantic part correspondence.}
      We report the average unsupervised keypoints correlation for each category. $\uparrow$ is better.
      Extended version with additional categories and detailed correlation tables can be found in the supplementary.
  }
  \label{t:seg}
\end{table}
  
\begin{figure}
\centering
\includegraphics[width=1\linewidth]{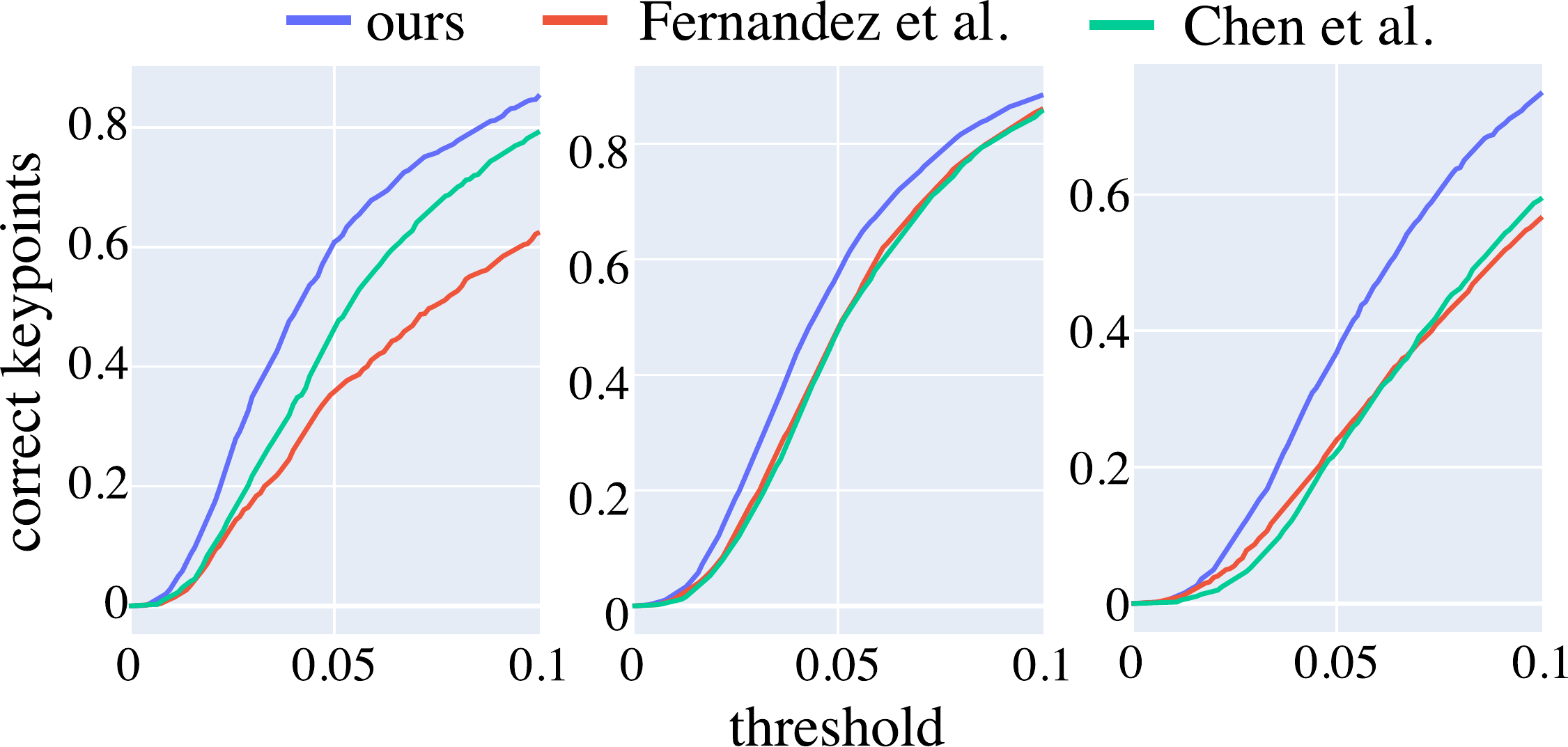}
\small
(a) airplane \hspace{14mm} (b) car \hspace{14mm} (c) chair

\smallskip
\smallskip

\setlength{\tabcolsep}{8pt}
\begin{tabular}{l|ccccc}
  \toprule
    & airplane & car & chair \\ \midrule
    Chen~\etal~\cite{chen2020unsupervised} & 0.49 & 0.46 & 0.22  \\
    Fernandez~\etal~\cite{fernandez2020unsupervised} & 0.36 & 0.47 & 0.24 \\
    ours & \textbf{0.61} & \textbf{0.56} & \textbf{0.37} \\
  \bottomrule
\end{tabular}
\smallskip

(d) Percentage of correct keypoints (PCK) @0.05

\caption{
    \textbf{Unsupervised 3D keypoints accuracy.}
    We measure the semantic consistency of keypoints following~\cite{thewlis2017unsupervised}.
    We train a linear regressor to predict manually annotated keypoints from unsupervised keypoints.
    The regressor accuracy on the test set estimates the semantic consistency of the underlying unsupervised keypoints.
    We show results in terms of PCK for airplane, car and chair category on the KeypointNet dataset~\cite{you2020keypointnet}.
}
\label{f:pck}
\end{figure}

\begin{figure}
\centering
\includegraphics[width=1\linewidth]{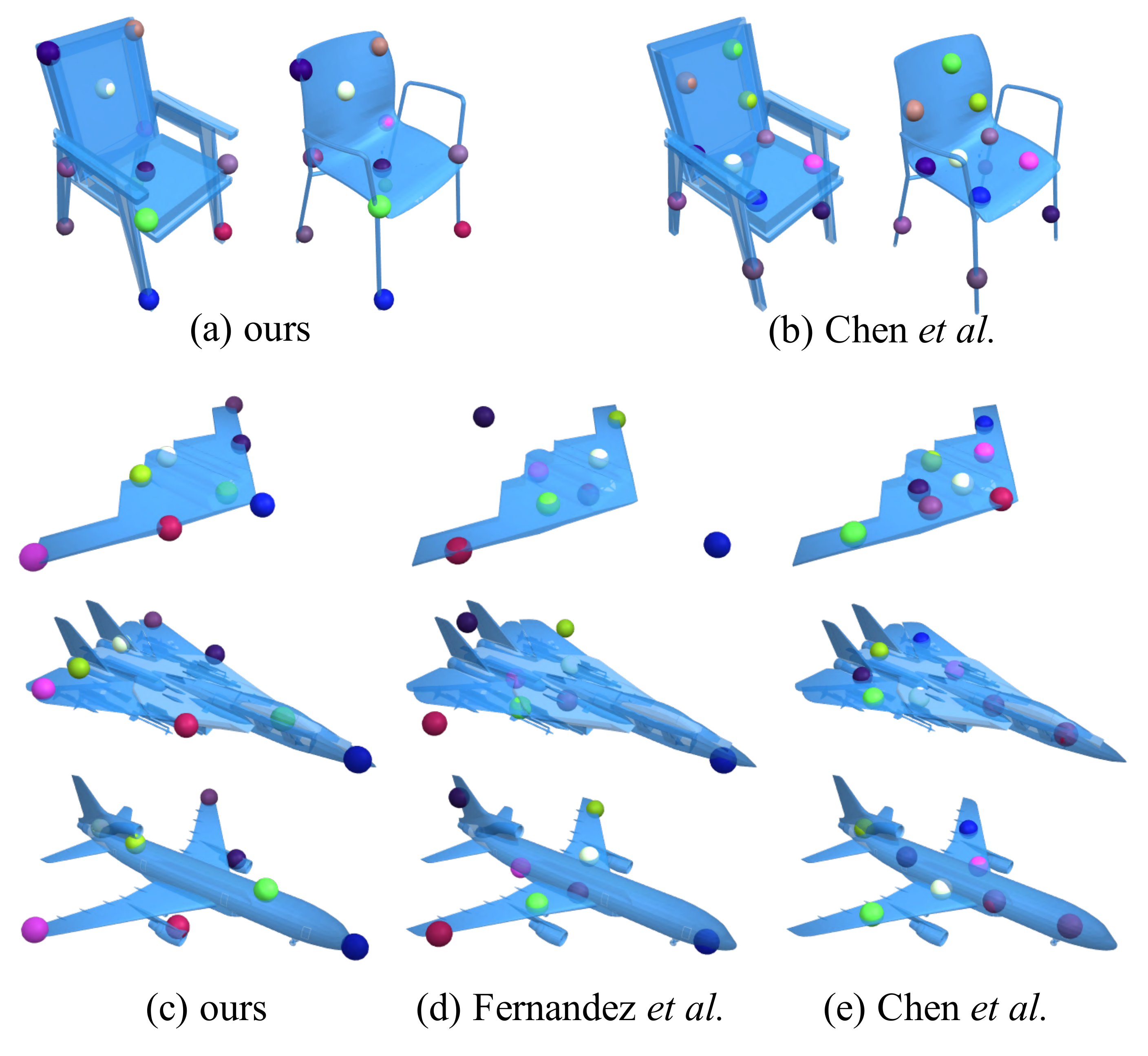}
\caption{
    \textbf{Unsupervised 3D keypoints.}
    We compare our unsupervised 3D keypoints with Fernandez \etal~\cite{fernandez2020unsupervised} and Chen \etal~\cite{chen2020unsupervised}.
    Our keypoints are more semantically consistent despite large shape variations when compared to other methods.
    Keypoints obtained by Fernandez \etal~\cite{fernandez2020unsupervised} do not explain all the shapes well.
    Moreover, our keypoints are symmetrical without explicitly enforcing that in contrast with~\cite{fernandez2020unsupervised}.
    We show results on additional categories in the supplementary.
}
\vspace{-1em}
\label{f:keypoints}
\end{figure}

\begin{figure}[]
  \small
  \centering
  \includegraphics[width=1\linewidth]{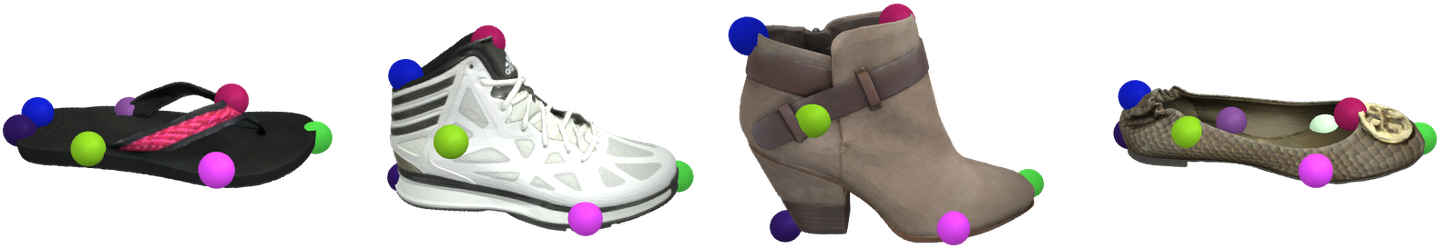}
  \includegraphics[width=1\linewidth]{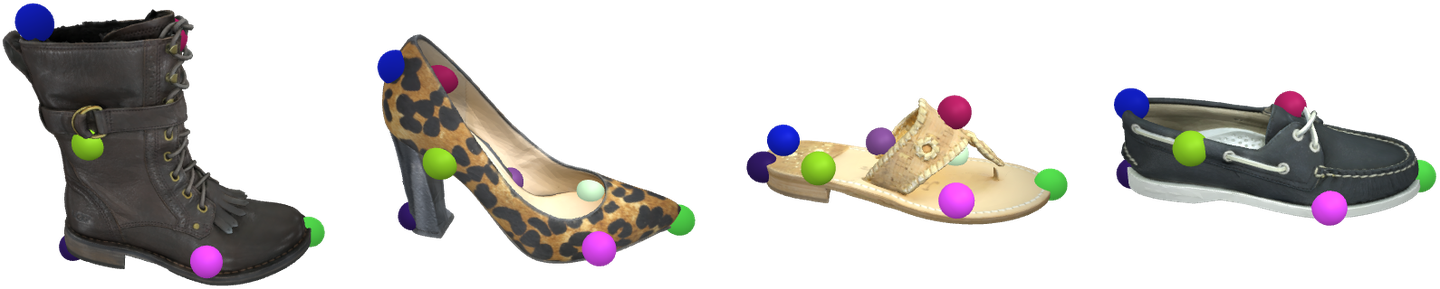}
  \caption{
      \textbf{Unsupervised 3D keypoints on real-world data.}
      We run our unsupervised keypoint detector on real-world scans of shoes~\cite{GoogleScannedObjects}.
      The keypoints are semantically consistent across different shapes.
  }
  \label{f:shoes}
  \vspace{-1em}
\end{figure}
  
\begin{figure}[]
  \small
  \centering
  \includegraphics[width=1\linewidth]{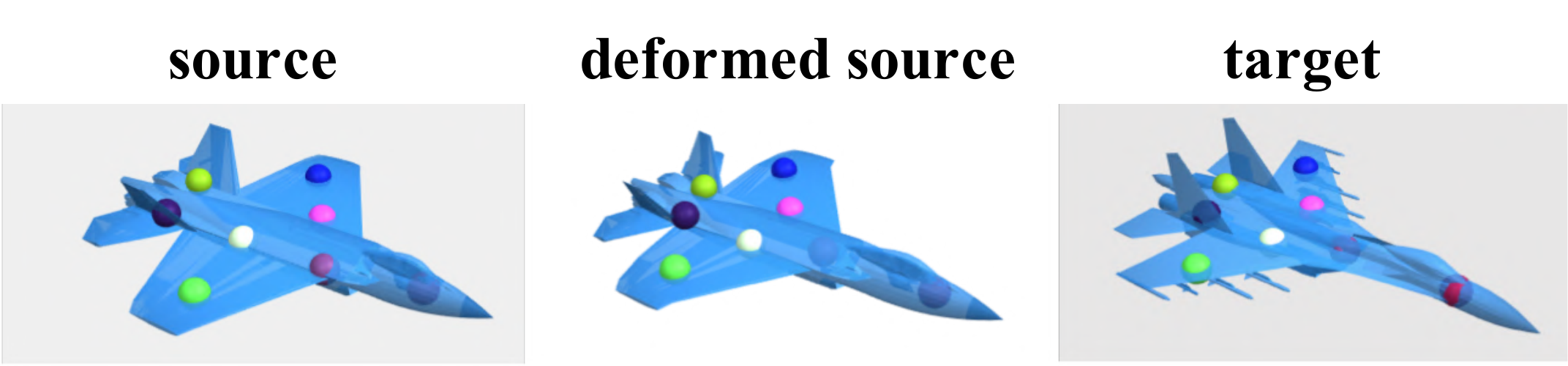}
  \vspace{2mm}
  (a) Chen \etal~\cite{chen2020unsupervised}
  \includegraphics[width=1\linewidth]{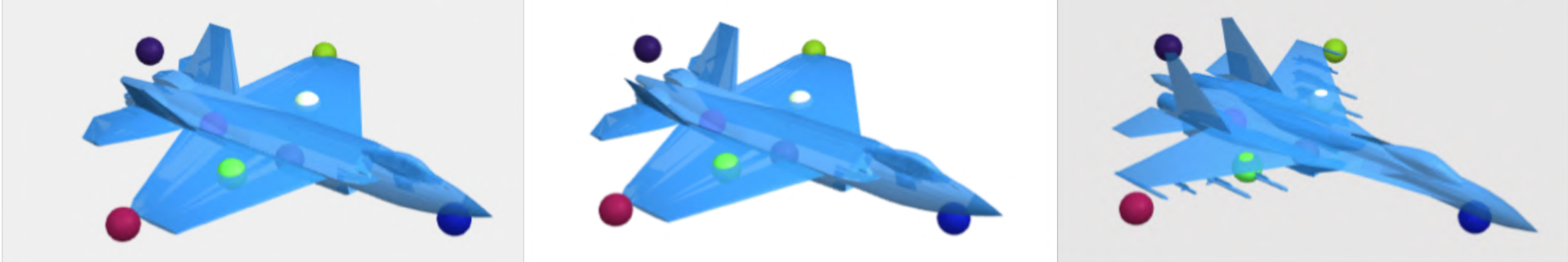}
  \vspace{2mm}
  (b) Fernandez \etal~\cite{fernandez2020unsupervised}
  \includegraphics[width=1\linewidth]{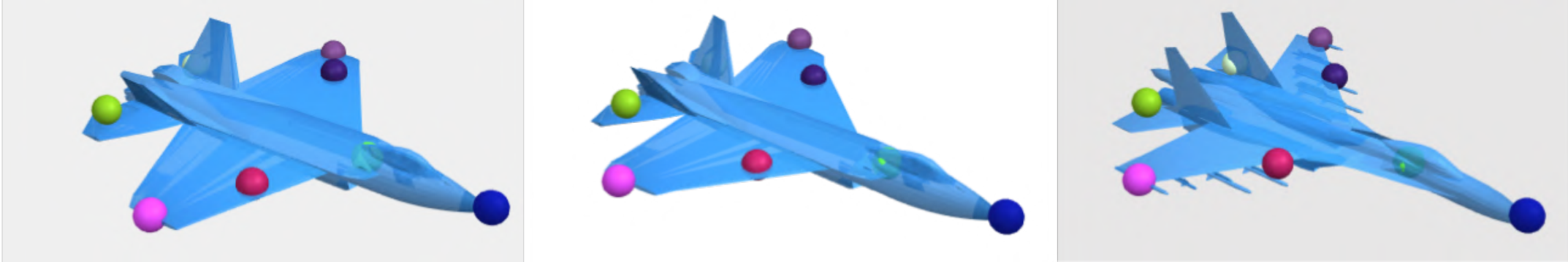}
  \vspace{2mm}
  (c) ours
  
  \resizebox{\linewidth}{!}{
  {
  \setlength{\tabcolsep}{2.8pt}
  \centering
  \begin{tabular}{l|cccc}
    \toprule
      & Fernandez~\etal~\cite{fernandez2020unsupervised} & Chen~\etal~\cite{chen2020unsupervised} & annotations~\cite{you2020keypointnet} & ours \\ \midrule
    CD & 7.55 & 5.93 & 4.20 & \textbf{3.02} \\
    \bottomrule
    \end{tabular}%
  }
  }
  \smallskip
  
  (d) Chamfer distance between deformed source and target
  \caption{
      \textbf{Keypoints for shape deformation.}
      We replace our discovered keypoints in KeypointDeformer to compare with different keypoints detectors and manually annotated keypoints on keypoint-guided pairwise shape alignment for the airplane category.
      The degree of alignment is measured by the Chamfer distance between the deformed source and target shapes.
      Our discovered keypoints can align shapes better even when compared to manually selected keypoints from KeypointNet~\cite{you2020keypointnet}.
      Keypoints from Fernandez \etal~\cite{fernandez2020unsupervised} and Chen \etal~\cite{chen2020unsupervised} fail to accurately align shapes as their keypoints are less precise.
      Data in the table are scaled by $10^3$.
  }
  \label{f:deform}
\end{figure}
  
\begin{figure}[]
  \small
  \centering
  \includegraphics[width=1\linewidth]{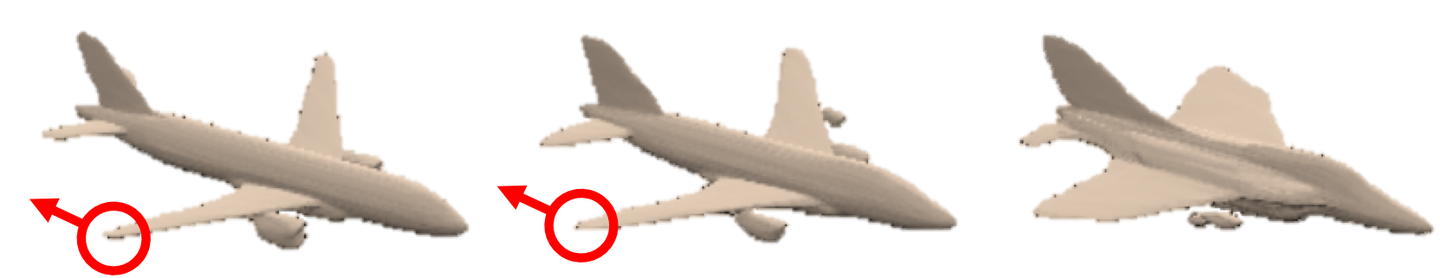}
  (a) DualSDF~\cite{hao2020dualsdf}
  \includegraphics[width=1\linewidth]{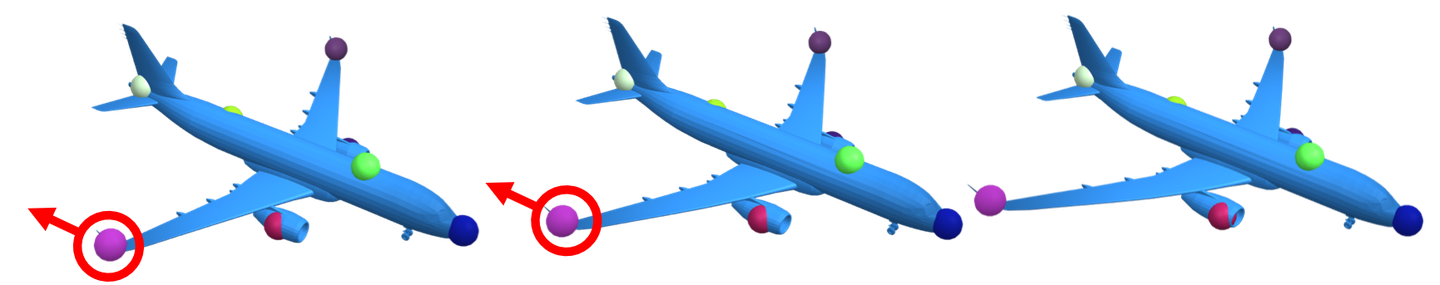}
  (b) ours
  \includegraphics[width=1\linewidth]{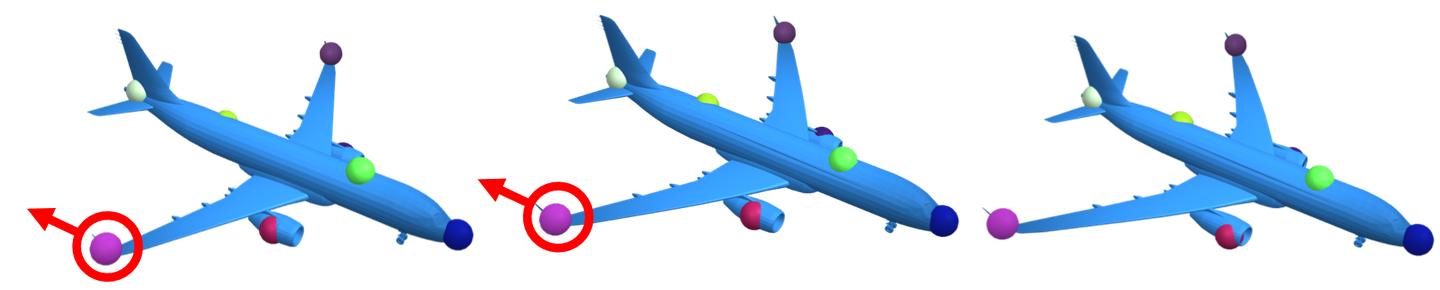}
  (c) ours with shape prior
  \caption{
      \textbf{Comparison with DualSDF~\cite{hao2020dualsdf}.}
      We move the wing tip in the direction of the red arrow.
      (a) DualSDF is a generative model and changing the position of the wing tip results in a change from an airliner to a jet fighter.
      In contrast, our method preserves the original structure of the mesh and allows for asymmetric manipulation when desired (b).
      Our method can also work in conjunction with a shape prior (\Cref{s:prior}) to achieve symmetrical manipulation (c).
  }
  \label{f:dualsdf}
  \vspace{-1em}
\end{figure}

\subsection{Experimental Setup}
\paragraph{Datasets.}
We train our KeypointDeformer using ShapeNet~\cite{chang2015shapenet} following the standard training and testing split.
We normalize all the shapes into a unit box.
For evaluation, we use semantic part annotations for ShapeNet~\cite{yi2016scalable}, as well as the KeypointNet~\cite{you2020keypointnet} dataset, which  contains semantic keypoint annotations for selected ShapeNet categories.
Note that our method does not require any of these annotations for training.
We also evaluate KeypointDeformer on real-world 3D scans of shoes from Google Scanned Objects dataset~\cite{GoogleScannedObjects}.

\medskip \noindent \textbf{Implementation details.}
The keypoint predictor $\Phi$ and the influence predictor $\Gamma$ are implemented as neural networks using a PointNet encoder and the whole model is optimized using the Adam optimizer.
We use 1024 sampled points for the point cloud representation of shape $\bx$.
Unless otherwise mentioned, we predict 12 unsupervised keypoints for all categories except for airplane and car where we use 8.
The supplementary contains an ablation studying the effect of different number of unsupervised keypoints.
We set the number of sampled farthest points $\bq$ to the double of the number of keypoints.
Detailed descriptions of network architectures and training details are in the supplemental material.

\subsection{Semantic Consistency}
\label{s:semantic}
We first demonstrate the quality of our unsupervised keypoints by evaluating their semantic consistency, \ie
whether they always correspond to the same semantic object parts or not.
For instance, 
if a keypoint is predicted on the tip of the wing on one instance of an airplane, then that same keypoint should always correspond to the tip of the wing across different instances. 
For this task we compare with recently introduced methods for unsupervised keypoint discovery from  Fernandez \etal~\cite{fernandez2020unsupervised} and Chen \etal~\cite{chen2020unsupervised}.

We evaluate semantic consistency using two protocols.
First, we use an evaluation protocol of Fernandez \etal~\cite{fernandez2020unsupervised}.
Since their evaluation is very coarse, we also follow an evaluation protocol for unsupervised keypoints established by Thewlis \etal~\cite{thewlis2017unsupervised}.

The evaluation protocol of Fernandez~\etal~\cite{fernandez2020unsupervised} employs
the ShapeNet dataset with part annotations to measure the correlation between each keypoint and annotated semantic object parts across instances of the category.
Each keypoint is associated with the nearest object part.
This protocol has two limitations.
First, a keypoint can be associated with an object part even if it lies far from the object (indicating a poor choice of keypoint).
Second, this protocol does not account for boundary keypoints that are predicted just between two annotated object parts (which can still be high-quality, salient keypoints).
To address these limitations, we propose a small modification to this protocol, in which we associate each keypoint with a given object part if it lies within its small neighborhood (0.05 from the object part when the object is normalized to unit box)---hence, a keypoint can be associated with multiple parts.
For each keypoint, we compute its correlation with each object part.
Since a keypoint can be associated with multiple parts, we consider only the most correlated part for a given keypoint in the final metric.
The final metric then computes the average correlation over all the keypoints.
We report semantic consistency results for ShapeNet categories in~\Cref{t:seg}.
Our keypoints show better average correlation when compared to Chen~\etal~\cite{chen2020unsupervised} and Fernandez \etal~\cite{fernandez2020unsupervised}.

Second, we adopt the standard unsupervised 2D keypoint evaluation protocol as in~\cite{thewlis2017unsupervised,zhang2018unsupervised,jakab2018unsupervised}, since the semantic object parts are coarsely annotated (e.g., the airplane category comes with only 3 semantic parts).
The objective of this protocol is to measure how predictive unsupervised keypoints are of semantic keypoints selected by humans.
This is done by finding a linear mapping between the unsupervised keypoints and manually annotated ones.
The linear mapping is established on the training set by fitting a linear regressor. %
The predictiveness of unsupervised keypoints is then measured in terms of this regressor's prediction error on the test set.
We use the recent KeypointNet dataset~\cite{you2020keypointnet}, which contains semantic annotations on ShapeNet dataset. %
We report the performance in~\Cref{f:pck}.
Our unsupervised keypoints are more predictive of manually annotated keypoints than other unsupervised keypoint.
\Cref{f:keypoints} provides qualitative comparison of our unsupervised keypoints with those obtained by other methods.

\paragraph{Real world scans.}
We also demonstrate applicability of our unsupervised keypoint detector on real-world 3D scans of objects.
We use the shoe category from Google Scanned Objects dataset~\cite{GoogleScannedObjects}.
We align the shapes using the automatical alignment method from~\cite{makadia06pami}.
We split the dataset into training and test sets with 219 and 36 samples respectively.
We use the same hyper-parameters as done in experiments on ShapeNet. 
\Cref{f:shoes} shows that our method learns semantically consistent 3D keypoints for shoes with largely different shapes.

\begin{figure*}
    \includegraphics[width=1\linewidth]{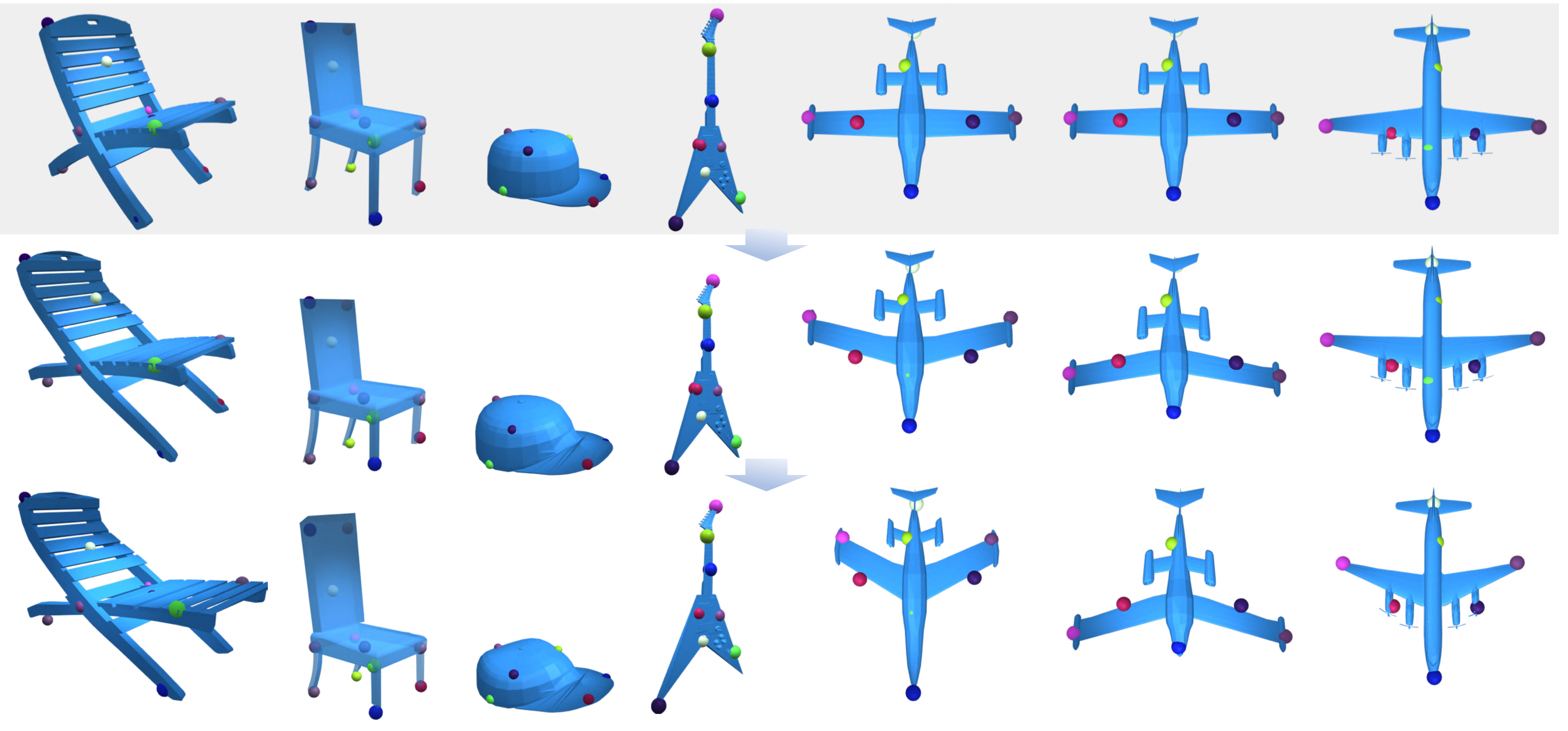}
    \caption{
    \textbf{Interactive shape control via 3D unsupervised keypoints.}
    We show iterative steps in user guided shape deformation using our discovered keypoint as handles.
    Top row shows initial state.
    Please refer to our project page for a demo video.
}
\label{f:control}
\end{figure*}

\subsection{Keypoints for Shape Deformation}\label{s:deform}
To quantitatively demonstrate that controlled shape deformation is possible through unsupervised keypoints, we use the task of pairwise shape alignment, in which we 
deform a source shape into a target shape.
In our case, the deformation is guided using keypoints. This task also evaluates that 
our discovered keypoints are more suitable for shape control than other keypoints. 
We modify our method by replacing our unsupervised keypoints with keypoints obtained from other methods.
We then train our deformation model from scratch.
We experiment with keypoints from~\cite{fernandez2020unsupervised}, \cite{chen2020unsupervised}, and also manually annotated keypoints from \cite{you2020keypointnet}.
Performance is evaluated by measuring the Chamfer distance between the deformed source shape and the target shape.
We present results in~\Cref{f:deform}.
The unsupervised keypoints obtained by other methods fail to capture the large variations in shapes in the dataset.
Our keypoints, on the other hand, can follow the large changes in shapes.
This ultimately leads to more accurate shape deformations.

\subsection{Shape Control via Unsupervised 3D Keypoints}
\label{s:control}

Our ultimate goal is to use automatically discovered keypoints to perform user-guided interactive shape deformation.
\Cref{f:control} shows interactive shape control using our unsupervised keypoints.
Our method provides low-dimensional handles to control object shape.
The control is intuitive as the deformation is semantically consistent, e.g., moving a keypoint on the leg of a chair or airplane wing results in movement of that object part in the same direction.
Thus the user can easily edit shape meshes.
Please refer to our project page for a demo video showcasing user-guided interactive shape control using keypoints.

The related work DualSDF~\cite{hao2020dualsdf} also allows for user-guided interactive shape deformation.
However, the key distinction here is that DualSDF is a conditional generative model.
Manipulating an object through its handle generates a new shape that respects the new position of the handle specified by the user,
but the new generated shape can be very different from the original one.
This aspect is illustrated in~\Cref{f:dualsdf}, where DualSDF transforms an airliner to a jet fighter.

\subsection{Categorical Shape Prior}\label{s:prior}
Since our deformation model uses keypoints as its low-dimensional shape representation, we can compute categorical shape prior on them. %
We compute PCA on the set of predicted keypoints obtained from the training set. We set the number of basis to 8.
As discussed in~\Cref{s:prior}, we use the prior in two ways.
First, we can use it during interactive shape control when the user manipulates only a single keypoint, to ``synchronize'' the rest of the keypoints according to the prior.
This ``synchronized'' editing is used in~\Cref{f:dualsdf} where we drag only a single keypoint and the rest get automatically readjusted.
\begin{figure}[t]
    \includegraphics[width=1\linewidth]{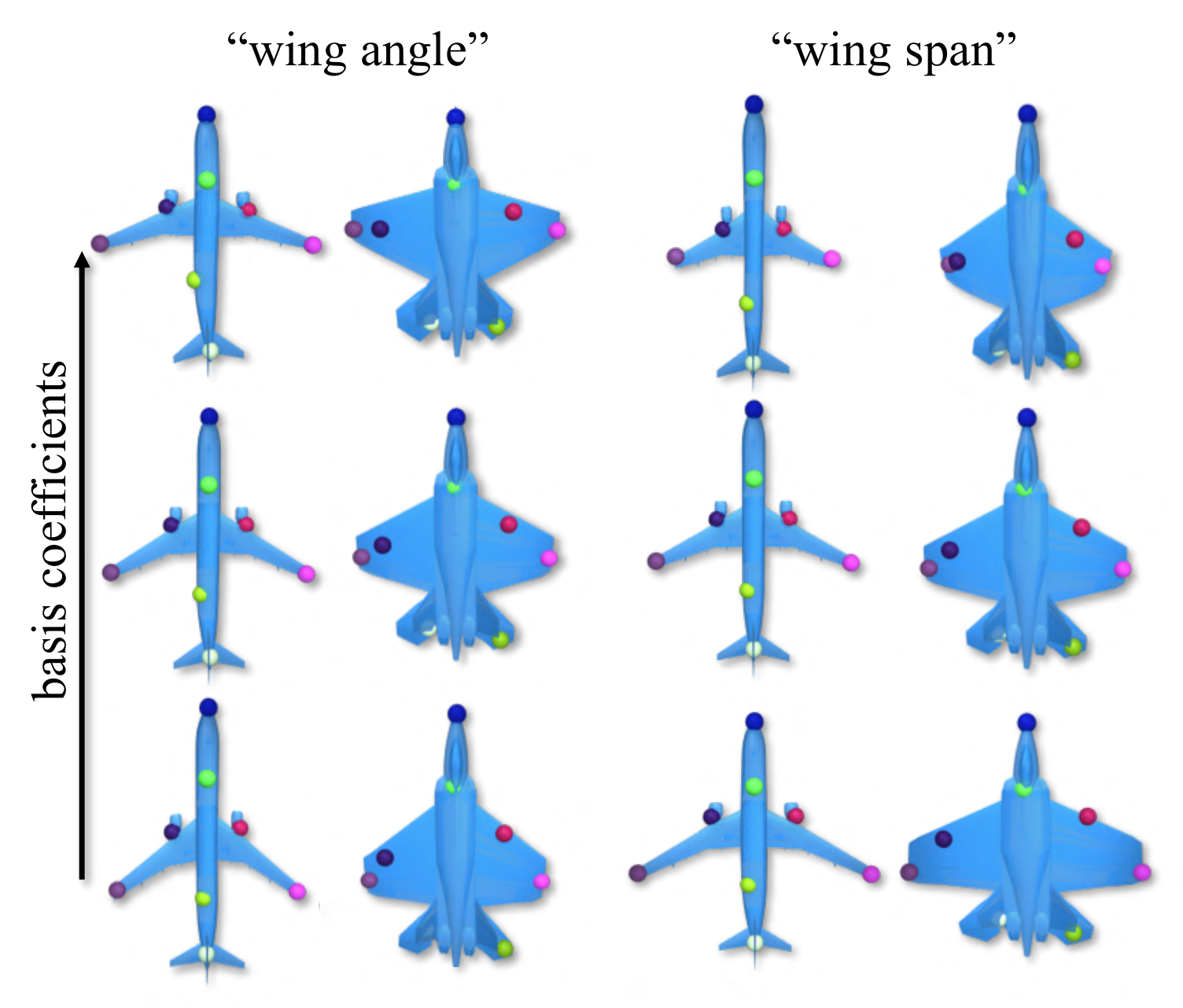}
\caption{
    \textbf{Varying PCA basis coefficients for shape augmentation.}
    We sample new keypoints by varying its PCA basis coefficients.
    The sampled keypoints are used to deform the original shape obtaining a new set of shapes.
    The left two columns show results for a subspace that correlates with the wing angle.
    The right two columns show results for a subspace that correlates with the wing span.
}
\label{f:pca}
\end{figure}

Second, we can easily sample new deformations using sampled keypoints that we obtain by varying PCA basis coefficients.
This can be applied to automatic dataset amplification as demonstrated~\Cref{f:pca}.

\section{Conclusion}
We present a method for controlling the shape of 3D objects through automatically discovered semantic 3D keypoints and a deformation model learned jointly with the keypoints. The resulting KeypointDeformer model provides users with a simple interface for interactive shape control. 
One limitation of the method is that our approach assumes aligned shape collections.
However, in our experiments with real scans, automatic alignment method was sufficient. 
Another limitation is that the keypoint representation does not allow modeling of individual object part rotations.
In this work we focused on the task of shape control and keypoint prediction, however 3D keypoints has various usage in other applications such as robotics~\cite{manuelli2019kpam,manuelli2020keypoints}. It would be interesting to explore the applicability of our unsupervised 3D keypoints to other tasks in the future. 

\newpage
{\small
\bibliographystyle{ieee_fullname}
\bibliography{refs}
}

\clearpage
\onecolumn
\section*{Appendix}
\appendix
This supplemental document provides an ablation study (\Cref{ss:abla}), extensive results (\Cref{ss:control,ss:keypoints}), and further implementation details (\Cref{ss:impl}). 
Please also refer to our video on our project page\footnotemark that demonstrates our method in action.

\footnotetext{\url{http://tomasjakab.github.io/KeypointDeformer}}

\section{Ablations}\label{ss:abla}
\begin{figure*}[h]
  \small
  \centering

  \begin{minipage}[b]{0.49\textwidth}
    \centering
    \includegraphics[width=\textwidth]{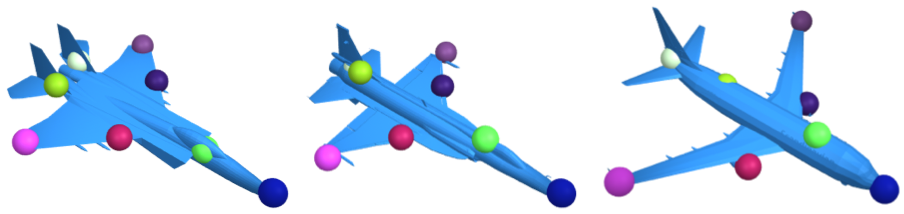}
    (a) 16 sampled farthest points
  \end{minipage}
  \hfill
  \begin{minipage}[b]{0.49\textwidth}
    \centering
    \includegraphics[width=\textwidth]{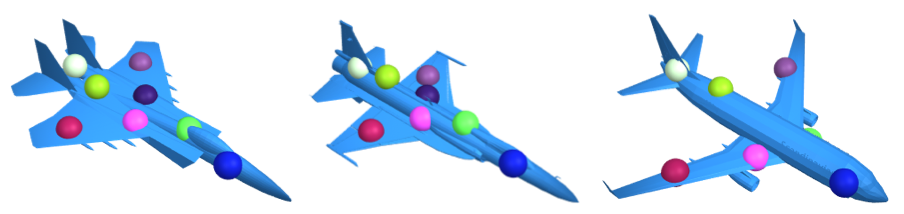}
    (b) 1024 points (full point cloud)
  \end{minipage}

  \begin{minipage}[b]{0.49\textwidth}
    \centering
    \includegraphics[width=\textwidth]{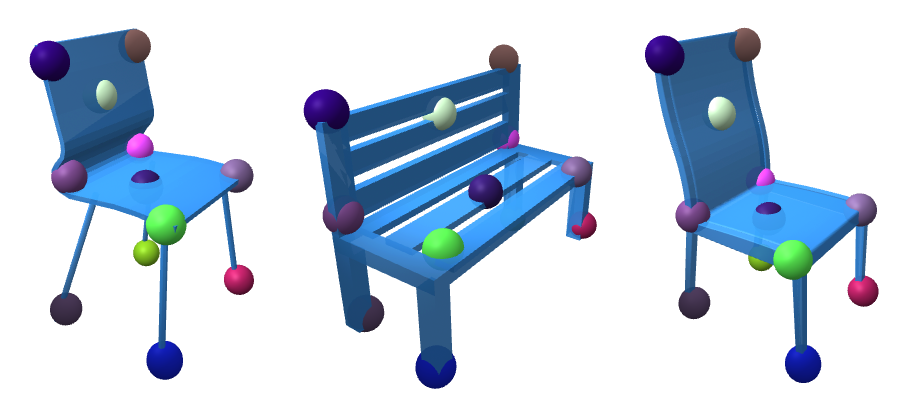}
    (c) 24 sampled farthest points
  \end{minipage}
  \hfill
  \begin{minipage}[b]{0.49\textwidth}
    \centering
    \includegraphics[width=\textwidth]{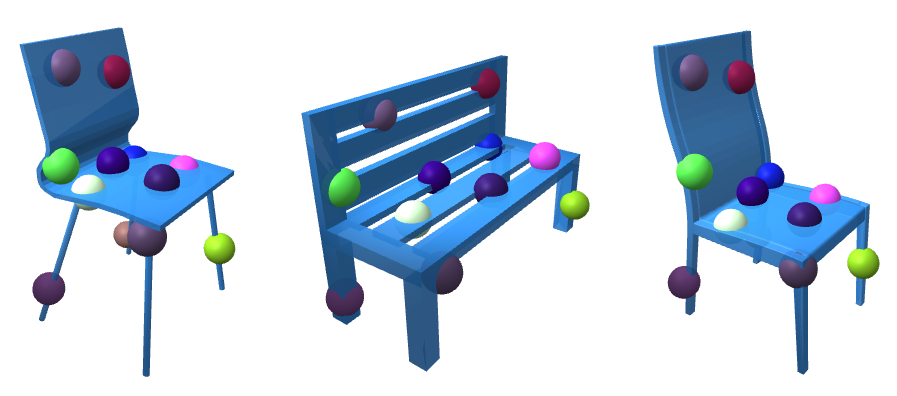}
    (d) 1024 points (full point cloud)
  \end{minipage}

  \smallskip
  \caption{
      \textbf{Farthest Point Keypoint regularizer ablation.}
      We investigate the influence of the number $J$ of sampled farthest points $\bq$ used for the keypoint regularizer (Section 3.2 of the main paper) on the quality of discovered unsupervised keypoints.
      We show unsupervised 3D keypoints trained using two versions of regularization.
      First, we set the number of sampled farthest points to double of the number of keypoints \textbf{(a, c)}.
      This is the setup that we use throughout the paper.
      Second, we set the number of sampled farthest points to the number of points in the point cloud representing the shape.
      This essentially results in a regularizer that is minimizing the Chamfer distance between the unsupervised keypoints and the object point cloud.
      Although the learned unsupervised keypoints have a good coverage \textbf{(b, d)} they are not as equally spaced and characteristic of the shape as \textbf{(a, c)}.
  }
  \label{st:fps}
\end{figure*}

\paragraph{Varying number of regularizing points.}
We examine the importance of the number of sampled farthest points $\bq$ on the quality of keypoint regularization (Section 3.2 of the main paper).
~\Cref{st:fps} shows the effect of different numbers of sampled farthest points on the discovered keypoints.
Using a high number of sampled farthest points in the regularization fails to learn keypoints that are equally spaced and characteristic of the underlying shape.

\begin{figure*}[t]
  \centering
  \includegraphics[page=1,width=\textwidth]{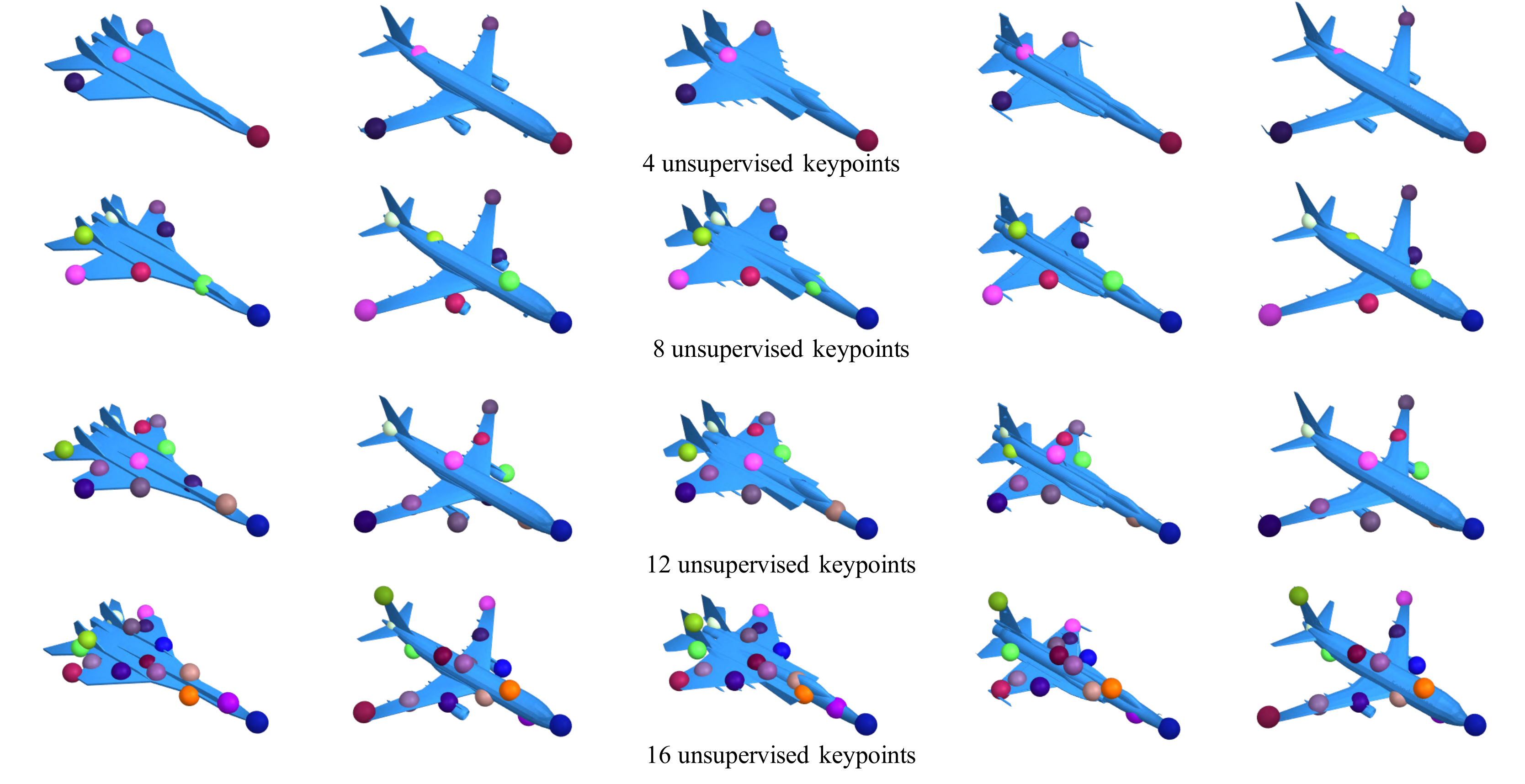}
  
  (a) predicted unsupervised keypoints
  \vspace{2mm}

  \begin{tabular}{l|cccc}
    \toprule
     \# of unsupervised keypoints & 4 & 8 & 12 & 16 \\ \midrule
     PCK@0.05 &  0.56 & 0.61 & 0.71 & 0.71 \\
    \bottomrule
  \end{tabular}

  \vspace{2mm}
  (b) quantitative evaluation

  \caption{
  \textbf{Varying number of keypoints.}
  The figure (a) shows the effect of different number of discovered keypoints (4, 8, 12, 16 from the top). The results are shown on randomly sampled results for ShapeNet Airplane category. Results in the table (b) are in terms of PCK@0.05 on airplanes from the KeypointNet dataset.
  }\label{sf:n_kp}
\end{figure*}

\paragraph{Varying number of keypoints.}
We vary the number of unsupervised keypoints discovered by our method.
\Cref{sf:n_kp} shows that our keypoints remain semantically consistent for different numbers of discovered keypoints.

\section{Shape Control via Unsupervised 3D Keypoints}\label{ss:control}
We show user guided interactive shape control in our supplementary video on our project page.
\Cref{sf:control} shows frames captured from user-guided interactive shape editing.
Editing using our keypoints is fast and intuitive while preserving the character and details of the original shape.

\section{Unsupervised 3D Keypoints}\label{ss:keypoints}
\begin{figure*}
  \small
  \centering
  \includegraphics[width=\textwidth]{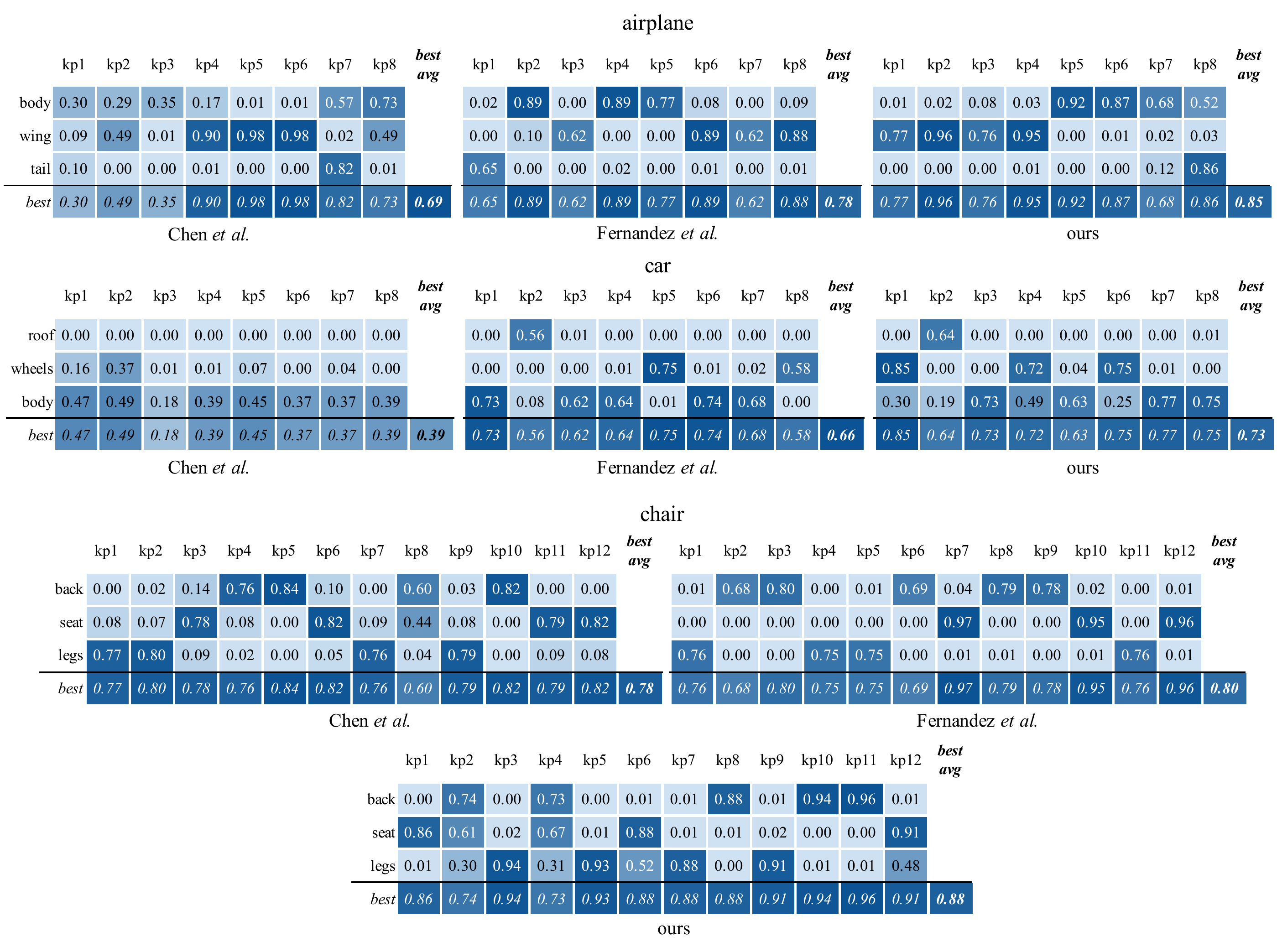}
  (a) unsupervised keypoints correlation
  \smallskip
  \smallskip

  \begin{minipage}{\textwidth}
    \footnotesize
      \setlength{\tabcolsep}{3pt}
      \resizebox{\linewidth}{!}{
      \begin{tabular}{l|cccccccccccccccc}
        \toprule
         & airplane & cap & car & chair & guitar & knife & laptop & motorbike & mug & skateboard & table  & pistol & bag & rocket & earphone & lamp \\ \midrule
          Chen~\etal~\cite{chen2020unsupervised} & 0.69 & 0.24 & 0.39 & 0.78 & 0.97 & 0.94 & 0.95 & 0.91 & 0.50 & 0.89 & 0.75 & 0.78 & 0.35 & 0.56 & 0.30 & 0.50 \\
          Fernandez~\etal~\cite{fernandez2020unsupervised} & 0.78 & 0.45 & 0.66 & 0.80 & 0.93 & 0.92 & 0.85 & 0.90 & 0.78 & 0.92 & 0.85 & 0.60 & 0.72 & 0.61 & 0.24 & 0.40 \\
          ours & \textbf{0.85} & \textbf{0.71} & \textbf{0.73} & \textbf{0.88} & \textbf{0.99} & \textbf{0.96} & \textbf{0.96} & \textbf{0.93} & \textbf{0.94} & \textbf{0.96} & \textbf{0.92} & \textbf{0.91} & \textbf{0.85} & \textbf{0.90} & \textbf{0.72} & \textbf{0.53} \\
        \bottomrule
      \end{tabular}}
  \end{minipage}

  \smallskip
  
  (b) average unsupervised keypoints correlation

  \caption{
  \textbf{Semantic part correspondence.}
  We evaluate semantic part correspondence for the ShapeNet Car category.
  The tables (a) shows the frequency of each unsupervised keypoint \textbf{[kp*]} being associated with a given object part.
  Chen \etal~\cite{chen2020unsupervised} show worse performance in this task because that methods tends to predict keypoints inside the object far from the annotated object surface.  We also report the average unsupervised keypoints correlation for each category (b). $\uparrow$ is better.
  }\label{sf:seg}
\end{figure*}

\begin{figure*}
  \centering
  \includegraphics[width=\textwidth]{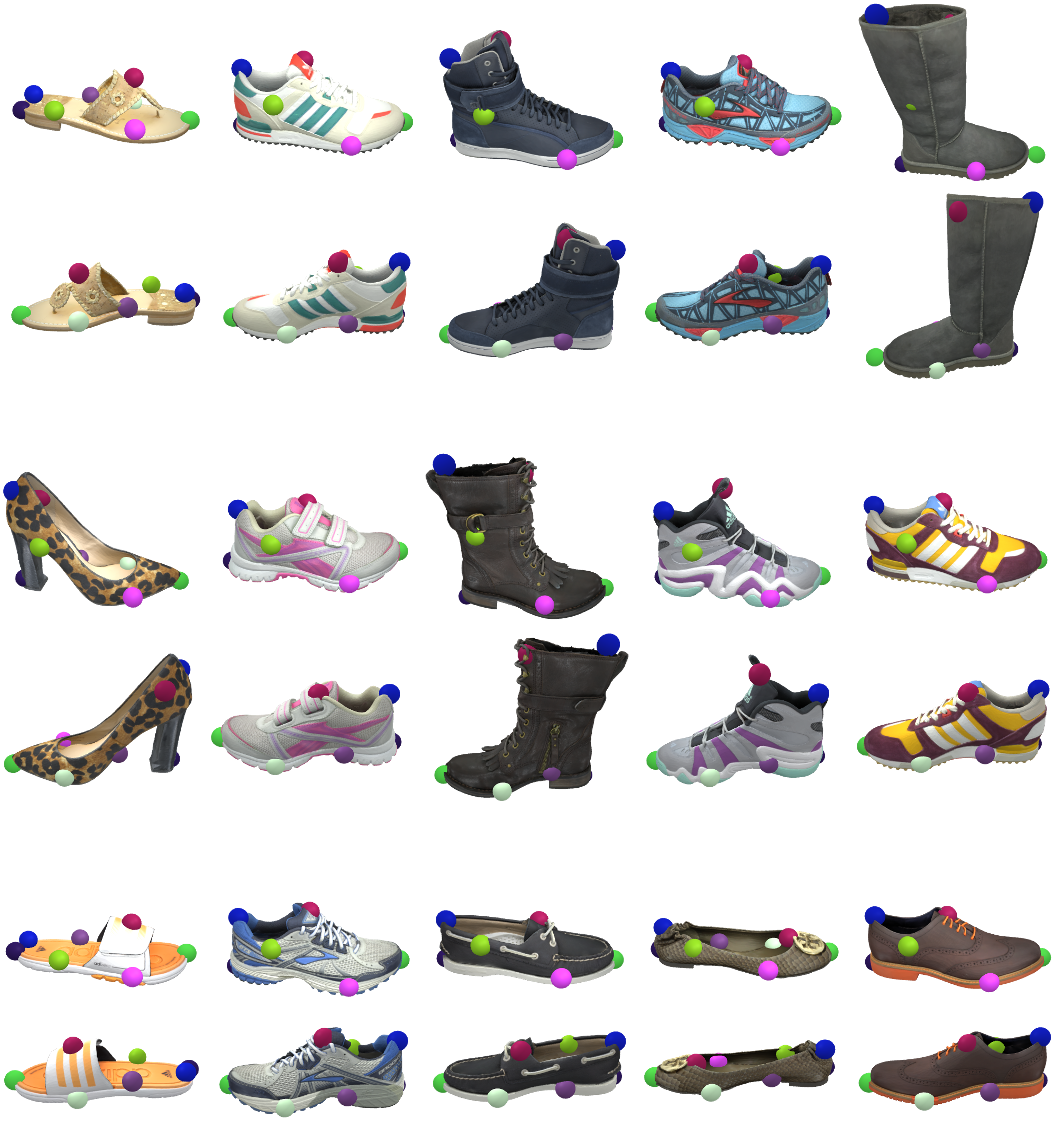}
  \caption{
  \textbf{Unsupervised 3D keypoints on real-world 3D scans.}
  Randomly sampled results with 8 unsupervised keypoints on real-world 3D scans of shoes from Google Scanned Objects dataset~\cite{GoogleScannedObjects}.
  }\label{sf:kp-shoes}
\end{figure*}

\begin{figure*}
  \centering
  \includegraphics[width=\textwidth]{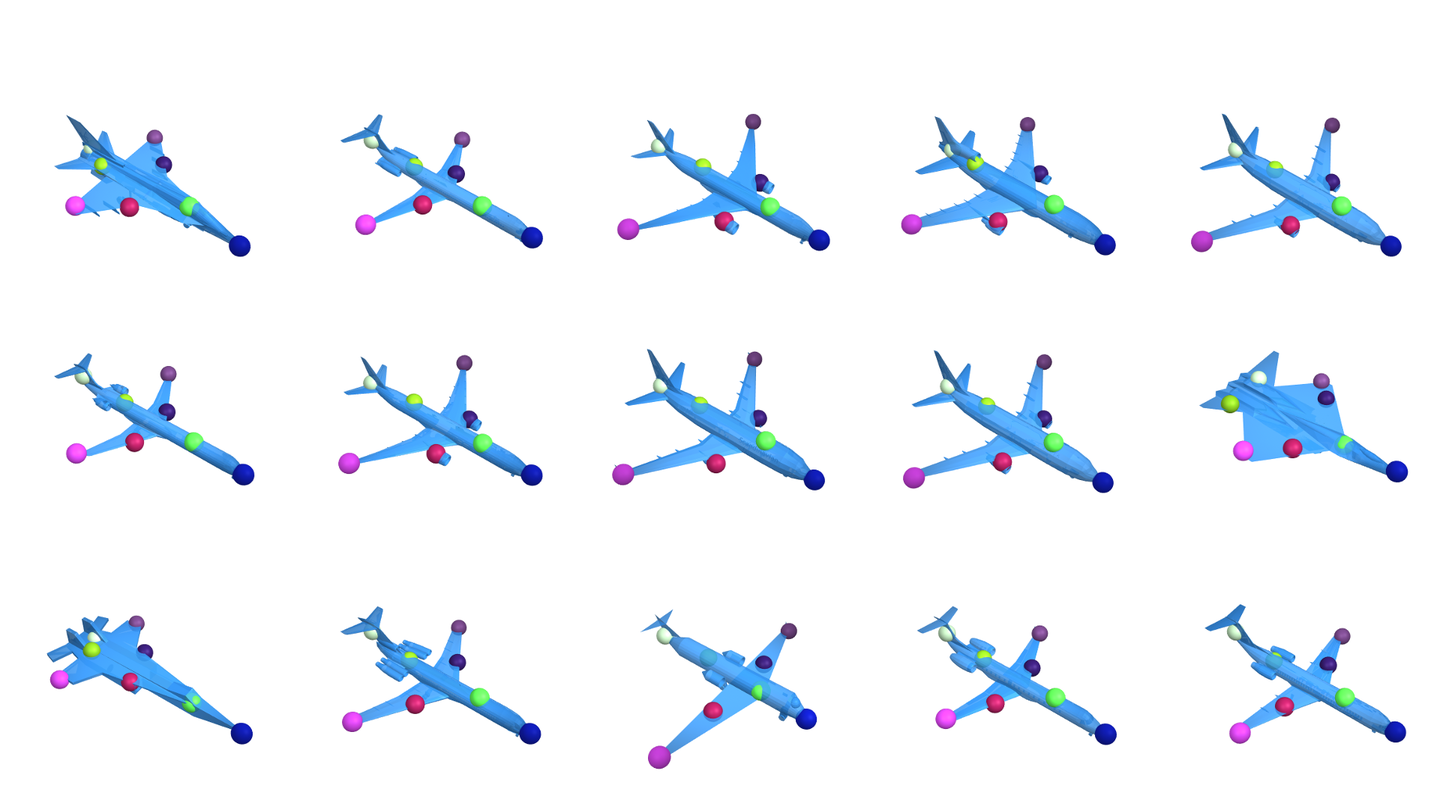}
  \caption{
  \textbf{Unsupervised 3D keypoints.}
  Randomly sampled results with 8 unsupervised keypoints for ShapeNet Airplane category.
  }\label{sf:kp-plane}
\end{figure*}

\begin{figure*}
  \centering
  \includegraphics[width=\textwidth]{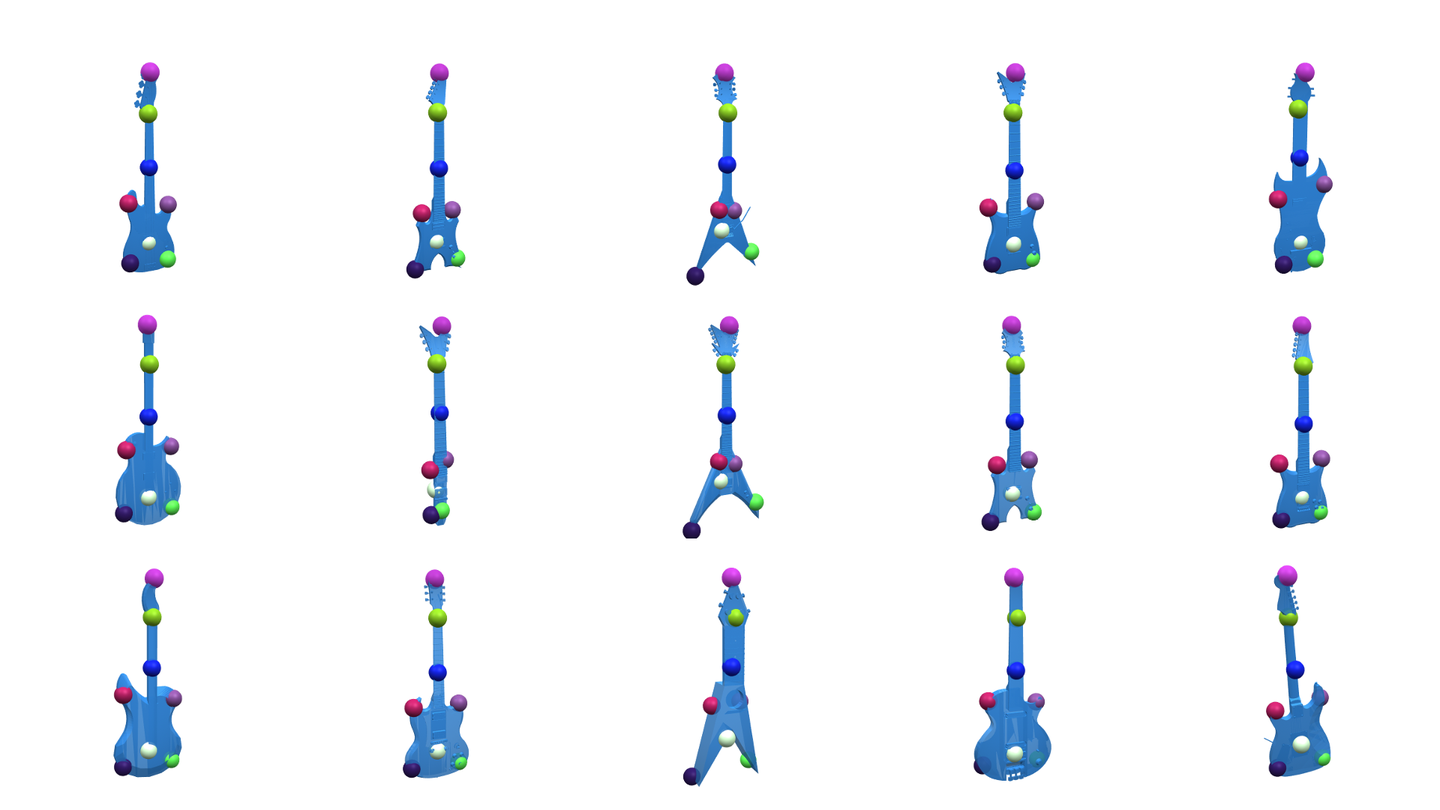}
  \caption{
  \textbf{Unsupervised 3D keypoints.}
  Randomly sampled results with 8 unsupervised keypoints for ShapeNet Guitar category.
  }\label{sf:kp-guitar}
\end{figure*}

\begin{figure*}
  \centering
  \includegraphics[width=\textwidth]{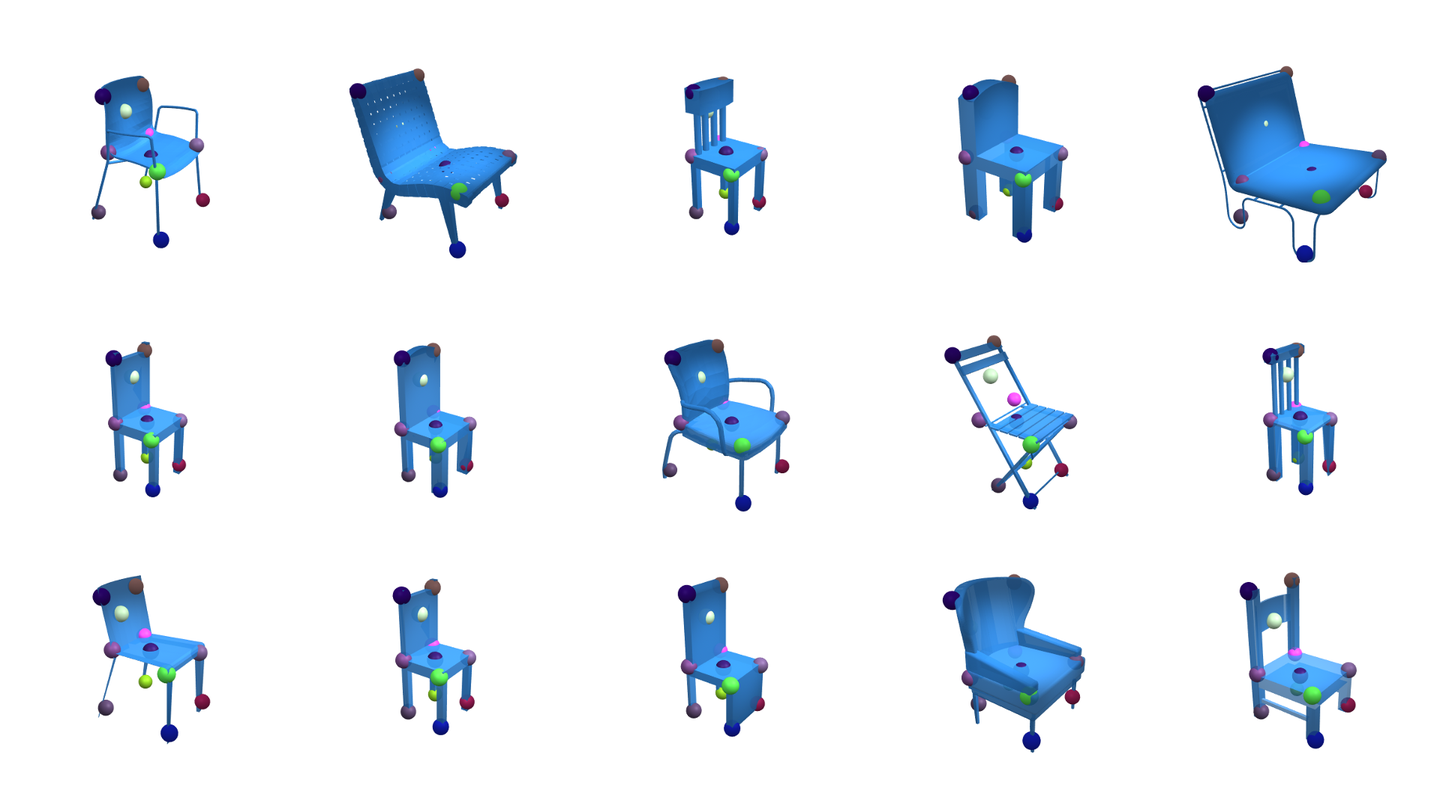}
  \caption{
  \textbf{Unsupervised 3D keypoints.}
  Randomly sampled results with 12 unsupervised keypoints for ShapeNet Chair category.
  }\label{sf:kp-chair}
\end{figure*}

\begin{figure*}
  \centering
  \includegraphics[width=\textwidth]{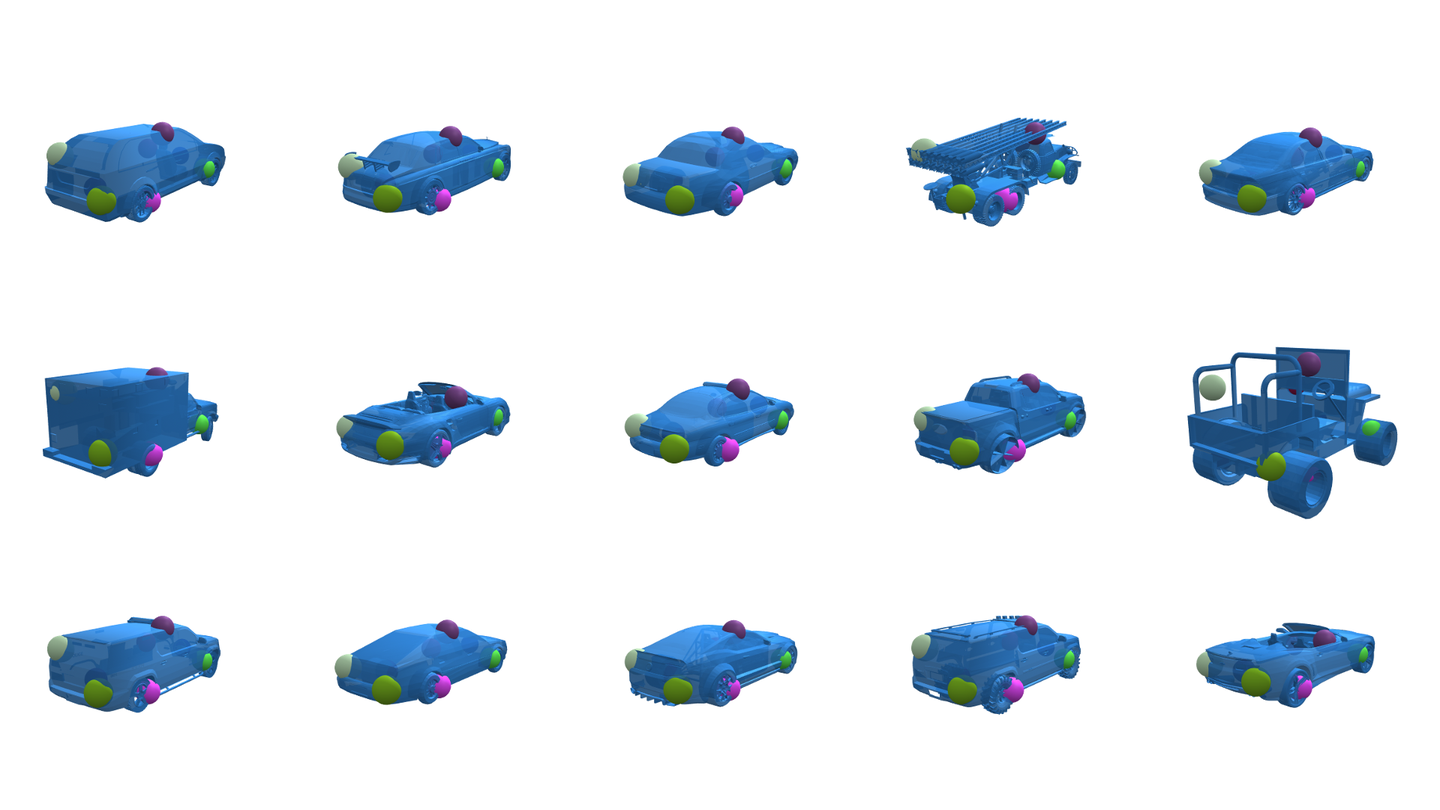}
  \caption{
  \textbf{Unsupervised 3D keypoints.}
  Randomly sampled results with 8 unsupervised keypoints for ShapeNet Car category.
  }\label{sf:kp-car}
\end{figure*}

We show extended quantitative results for semantic part correspondence experiment and detailed correlation tables in~\Cref{sf:seg} for the ShapeNet Car category.
\Cref{sf:kp-shoes,sf:kp-plane,sf:kp-guitar,sf:kp-chair,sf:kp-car} show extensive \emph{randomly} sampled qualitative test results for our unsupervised 3D keypoints.

\section{Implementation Details}\label{ss:impl}
\begin{figure*}
  \small
  \centering
  \begin{minipage}[b]{0.49\textwidth}
    \centering
    \includegraphics[width=\textwidth]{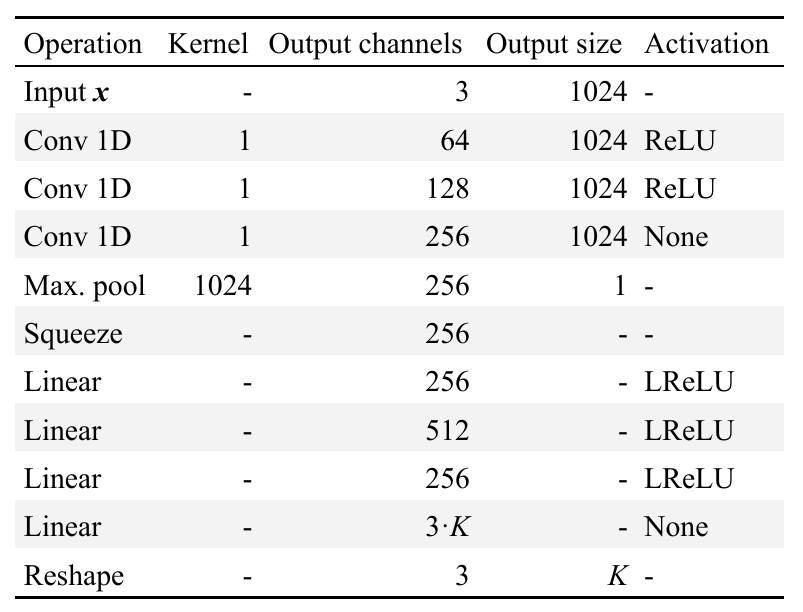}
    (a) Keypoint predictor $\Phi$
  \end{minipage}
  \hfill
  \begin{minipage}[b]{0.49\textwidth}
    \centering
    \includegraphics[width=\textwidth]{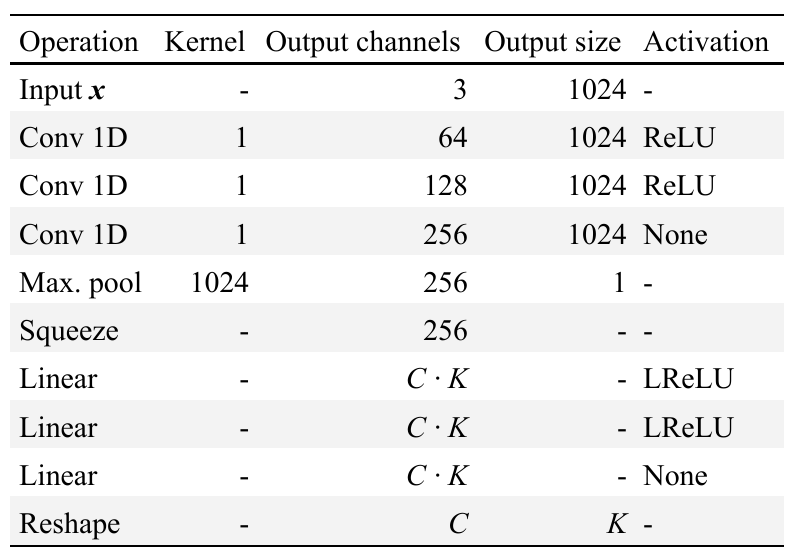}
    (b) Influence predictor $\Gamma$
  \end{minipage}
  \caption{\textbf{Network architectures.} 
  The network architectures are based on a PointNet encoder~\cite{qi2017pointnet,yifan2020neural}. 
  $K$ is the number of discovered keypoints, $C$ is the number of cage vertices.
  LReLU stands for Leaky ReLU with $0.1$ negative slope.
  }\label{sf:arch}
\end{figure*}

Our model assumes that the shapes are aligned (in the same orientation).
The initial cage is a 42-vertex icosphere.
We limit the influence matrix $W$ to influence at most $M$ nearest cages vertices (Section 3.1) per each keypoint, with $M = \lfloor C / K \rfloor$, where $C$ is the number of cage vertices and $K$ is the number of discovered keypoints.
We use a learning rate of 0.001.
The scalar loss coefficients (Section 3.2) $\alpha_\text{kpt}$ and $\alpha_\text{inf}$ are set to $1.0$ and $10^{-6}$ respectively.
\Cref{sf:arch} shows detailed description of network architectures used for the keypoint predictor $\Phi$ and the influence predictor $\Gamma$.

\paragraph{Datasets.}
KeypointNet~\cite{you2020keypointnet} dataset contains semantic 3D keypoint annotations for ShapeNet dataset~\cite{chang2015shapenet}.
Some models in KeypointNet are missing full keypoint annotations, therefore we use a subset of annotated keypoints that are contained in at least 80\% of the models.
KeypointNet also does not follow the standard training and testing splits from ShapeNet.
We resample the KeypointNet dataset splits to make it compatible with the original ShapeNet splits.

\begin{figure*}
    \includegraphics[width=1\linewidth]{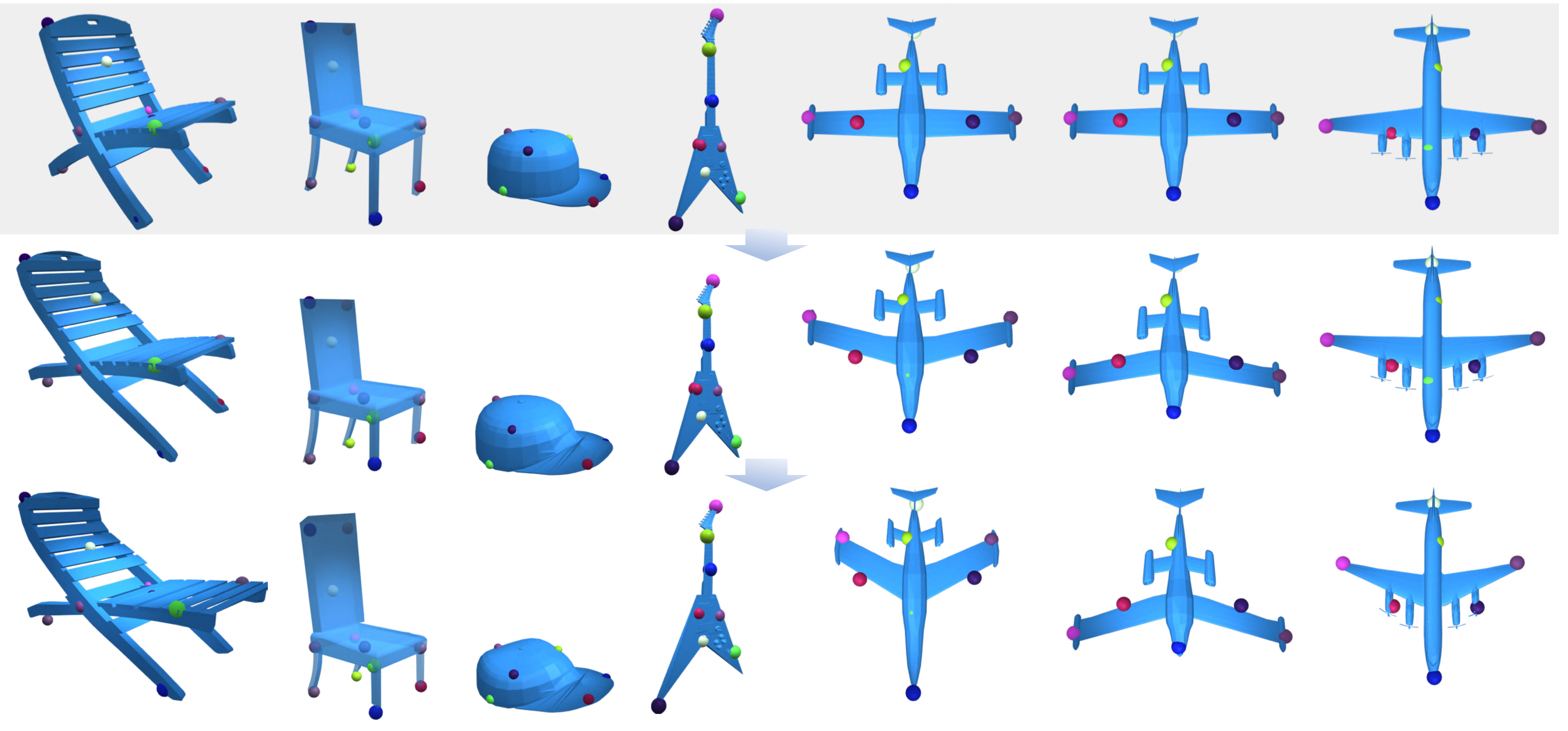}
    \caption{
    \textbf{Interactive shape control via 3D unsupervised keypoints.}
    We show iterative steps in user guided shape deformation using our discovered keypoint as handles.
    Top row shows initial state.
    Please refer to our project page for a demo video.
}
\label{f:control}
\end{figure*}

\clearpage
\twocolumn

\end{document}